\documentclass[journal]{IEEEtran}

\usepackage{algorithm}
\usepackage{algpseudocode}

\usepackage{float}
\usepackage{ifpdf}
\usepackage{cite}
\usepackage[pdftex]{graphicx}
\usepackage{amsmath,amssymb,amsfonts}
\usepackage{array}
\usepackage{multirow}
\usepackage{multicol}
\usepackage{textcomp}
\usepackage{colortbl}
\usepackage{booktabs}
\usepackage[table]{xcolor}
\usepackage{bm}
\usepackage{makecell}
\usepackage{longtable}
\usepackage[normalem]{ulem}
\useunder{\uline}{\ul}{}
\ifCLASSINFOpdf
\else
\fi

\begin{document}

\title{LightQANet: Quantized and Adaptive Feature Learning for Low-Light Image Enhancement}

\author{Xu Wu, Zhihui Lai$^*$, Xianxu Hou, Jie Zhou, Ya-nan Zhang, Linlin Shen

\thanks{$^*$ represents the corresponding author.} 

\thanks{Xu Wu, is with the College of Computer Science and Software Engineering, Shenzhen University, Shenzhen 518060, China and College of Computing and Data Science, Nanyang Technological University, Singapore (e-mail: csxunwu@gmail.com)}

\thanks{Zhihui Lai and Linlin Shen are with the Computer Vision Institute, College of Computer Science and Software Engineering, Shenzhen University, Shenzhen 518060, China. (e-mail: lai\_zhi\_hui@163.com}

\thanks{Xianxu Hou is with School of AI and Advanced Computing, Xi’an Jiaotong-Liverpool University, China (e-mail: hxianxu@gmail.com).}

\thanks{Jie Zhou is with the School of Mathematics and Statistics, Changsha University of Science and Technology, Changsha 410114, China, and also with the School of Artificial Intelligence, Shenzhen University, Shenzhen 518060, China. (e-mail: jie\_jpu@163.com).}

\thanks{Ya-nan Zhang is with School of Computer Science, Sichuan Normal University, Chengdu 610065, China (e-mail: 20240074@sicnu.edu.cn).}

\thanks{Linlin Shen is with Computer Vision Institute, School of Artificial Intelligence, Shenzhen University, Shenzhen 518060, China and also with the Department of Computer Science, University of Nottingham Ningbo China, Ningbo 315100, China (e-mail: llshen@szu.edu.cn).}
}

\markboth{Journal of \LaTeX\ Class Files,~Vol.~14, No.~8, August~2015}%
{Shell \MakeLowercase{\textit{et al.}}: Bare Demo of IEEEtran.cls for IEEE Journals}

\maketitle

\begin{abstract}
Low-light image enhancement (LLIE) aims to improve illumination while preserving high-quality color and texture. However, existing methods often fail to extract reliable feature representations due to severely degraded pixel-level information under low-light conditions, resulting in poor texture restoration, color inconsistency, and artifact.
To address these challenges, we propose LightQANet, a novel framework that introduces quantized and adaptive feature learning for low-light enhancement, aiming to achieve consistent and robust image quality across diverse lighting conditions.
From the static modeling perspective, we design a Light Quantization Module (LQM) to explicitly extract and quantify illumination-related factors from image features. By enforcing structured light factor learning, LQM enhances the extraction of light-invariant representations and mitigates feature inconsistency across varying illumination levels.
From the dynamic adaptation perspective, we introduce a Light-Aware Prompt Module (LAPM), which encodes illumination priors into learnable prompts to dynamically guide the feature learning process. LAPM enables the model to flexibly adapt to complex and continuously changing lighting conditions, further improving image enhancement.
Extensive experiments on multiple low-light datasets demonstrate that our method achieves state-of-the-art performance, delivering superior qualitative and quantitative results across various challenging lighting scenarios.

\end{abstract}


\IEEEpeerreviewmaketitle

\section{Introduction}
\IEEEPARstart{I}{mages} captured in dark environments, often referred to as low-light images, suffer from reduced illumination, increased artifact, and poor texture and color fidelity than those captured under normal-light conditions \cite{LLNet}. 
These deficiencies not only make it challenging for the human eye to discern objects but also significantly degrade the performance of advanced visual models, such as object detection systems. Therefore, the development of robust and effective LLIE methods is essential for improving the visual quality and utility of images captured in low-light conditions.

\begin{figure}[!t]
    \centering
    \includegraphics[width=0.45 \textwidth]{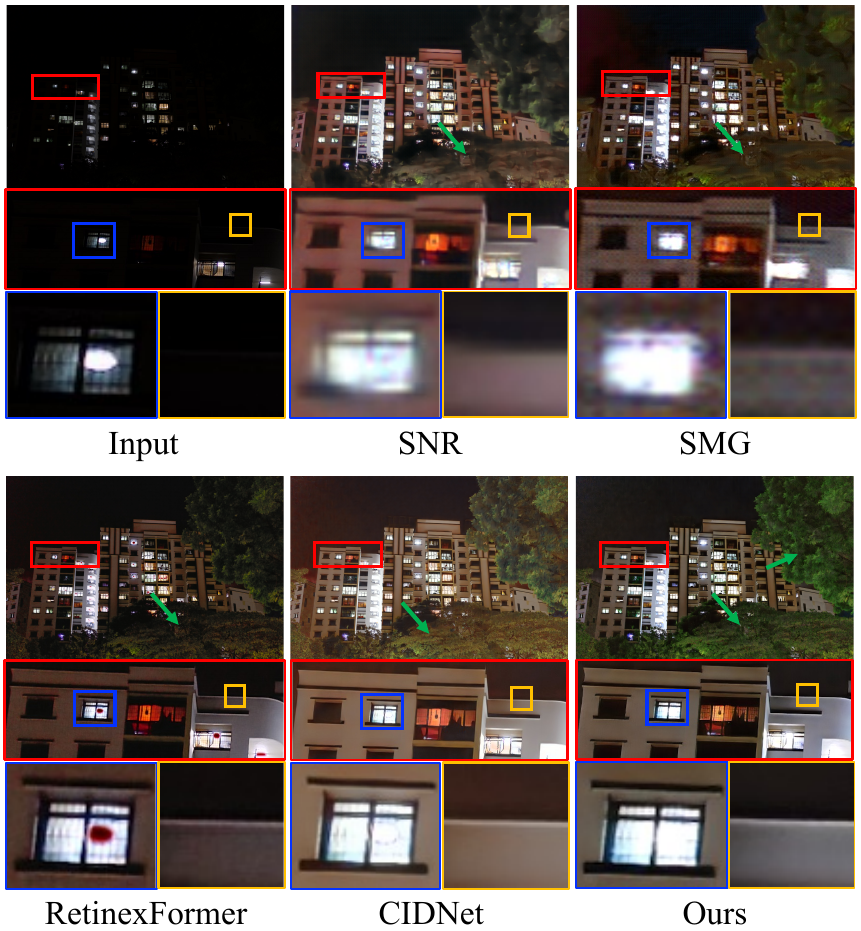}
    \caption{
    Visual comparisons on the real low-light image. The input image is compared with results from SNR\cite{SNR}, SMG\cite{SMG}, RetinexFormer\cite{Retinexformer}, CIDNet\cite{CIDNet}, and our method. The zoomed-in regions show the details of texture and sharpness restoration. Our method produces the most natural textures and smooth transitions around edges.
    }
    \vspace{-2mm}
    \label{fig:first_order}
\end{figure}

In previous works, Histogram Equalization (HE)-based and Retinex-based methods have been prominent in enhancing low-light images. HE-based methods enhance image contrast by adjusting the gray-level distribution of pixels to equalize the histogram \cite{AHE}. Conversely, Retinex-based methods focus on estimating and enhancing the illumination component of each pixel to improve brightness. However, both approaches may amplify noise and cause color distortion \cite{MEF}, presenting significant challenges that necessitate further refinement.

\begin{figure*}[!t]
    \centering
    \includegraphics[width=0.95 \textwidth]{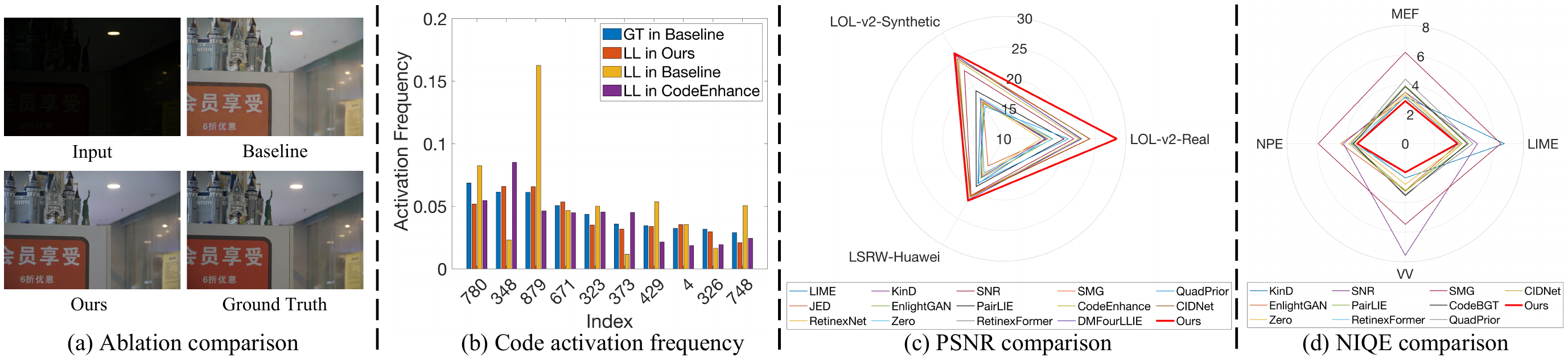}
    \vspace{-2mm}
    \caption{Effectiveness analysis and benchmark comparison. 
    (a) represents results of the baseline built by VQ-GAN \cite{VQGAN} and our method.
    (b) shows code activation frequency on LOL-v2 Real dataset \cite{LOLv2}. “GT in Baseline” and “LL in Baseline” represent inputting ground truth and low-light images to pre-trained VQ-GAN \cite{VQGAN} and fine-tuned VQ-GAN with low-light images, respectively. “LL in CodeEnhance” and “LL in Ours” denote inputting low-light images to CodeEnhance \cite{xu24} and our method.
    Activation frequency of “LL in Ours” is closer to that of “GT in Baseline”, which indicates our method can learn and activate important features under low-light conditions.
    (c) and (d) present performance comparison on LLIE datasets (LOL-v2-Synthetic \cite{LOLv2}, LOL-v2-Real \cite{LOLv2}, LSRW-Huawei \cite{LSRW}, LIME \cite{LIME}, MEF \cite{MEF}, VV \ref{vv}, and NPE \cite{NPE}) in terms of PSNR (the higher the better) and NIQE (the lower the better). 
    As highlighted by the bold red line, the proposed method consistently achieves the best demonstrating its superior.
    }
    \label{fig:codebook}
    \vspace{-4mm}
\end{figure*}

Recent advances in deep learning for LLIE have primarily focused on end-to-end networks that directly improve illumination \cite{LACR}.
Many of these methods introduce explicit illumination modeling, such as using an illumination branch to guide feature learning \cite{SNR}, or applying Retinex theory to decompose images into reflection and illumination components \cite{Retinexformer}.
However, despite these innovations, current approaches often struggle to preserve robust feature representations under low-light conditions, where severely compromised pixel-level visibility and reliability hinder effective feature extraction, ultimately resulting in texture degradation and color distortion.
As shown in Fig.~\ref{fig:first_order}, the results of SNR \cite{SNR} and SMG \cite{SMG} suffer from noticeable blurring and artifact amplification, leading to degraded texture clarity and unnatural visual appearances. RetinexFormer tends to cause over-exposure in brighter regions while under-enhancing darker areas, and CIDNet exhibits evident color distortion. By contrast, our method achieves clearer edge contours and more faithful texture recovery, producing sharper window patterns and smoother surfaces that better preserve the perceptual realism of the scene.

Addressing these challenges requires moving beyond direct pixel-level enhancement towards feature-level robustness under complex and variable illumination.
In this study, we reformulate the conventional image-to-image enhancement pipeline into an image-to-feature learning framework, which reduces the uncertainty inherent in the enhancement process by focusing on more abstract and stable feature representations.
Specifically, we employ a Vector-Quantized Generative Adversarial Network (VQ-GAN) \cite{VQGAN} to reconstruct high-quality images by leveraging vector-quantized features and learning an effective mapping between images and these features for LLIE tasks.
However, as illustrated in Fig.~\ref{fig:codebook} (a) and (b), directly applying VQ-GAN to LLIE leads to suboptimal results. Under low-light conditions, the activation frequency of “LL in Baseline”, where the baseline model is VQ-GAN, exhibits significant inconsistency compared to “GT in Baseline”, particularly at crucial feature indices. 
This is primarily due to VQ-GAN's lack of illumination-aware mechanisms and its reliance on clean feature distributions for effective codebook matching. Under extreme darkness, encoder features become misaligned with the learned codebook, resulting in color distortion and overexposure. 
These observations highlight the necessity of learning light-invariant representations and ensure effective feature quantization under low-light conditions.

To tackle these challenges, we propose LightQANet, a novel framework for low-light image enhancement based on quantized and adaptive feature learning.
The core of LightQANet is the Light Quantization Module (LQM), which aims to explicitly quantify illumination-related information embedded in image features. 
Instead of treating illumination variations implicitly, LQM is designed to learn a structured representation of lighting conditions by extracting and quantizing the so-called \textit{light factors} from both low-light (LL) and normal-light (NL) images.
To achieve this, LQM is trained to distinguish illumination levels through a supervised contrastive objective, enabling it to accurately capture and quantify the intensity and distribution of illumination in the feature space.
Once LQM acquires the ability to model illumination variations, it serves as an auxiliary guidance mechanism for the LightQANet. Specifically, the LightQANet is encouraged to minimize the feature-level discrepancies between different illumination conditions, thereby promoting the extraction of light-invariant representations.
Through this collaboration, LQM not only provides structured illumination supervision but also enhances the robustness and generalization of the overall enhancement framework. Note that LQM is not required during testing.

While LQM provides a structured quantization of illumination information, real-world lighting conditions are often complex and dynamically changing. 
To address this challenge, we introduce the Light-Aware Prompt Module (LAPM), which dynamically guides feature learning based on illumination priors.
Specifically, LAPM encodes illumination information into a set of learnable prompts, each capturing discriminative characteristics associated with different illumination levels. These prompts are adaptively fused with intermediate feature representations in the primary LLIE network, enabling the model to systematically adjust its feature learning process according to the estimated lighting conditions.
By dynamically injecting illumination-specific cues into the feature space, LAPM enhances the model's ability to generalize across a wide range of lighting environments.

The main contributions of this work are as follows: 
\begin{itemize}

    \item We propose LightQANet, a novel framework that performs quantized and adaptive feature learning to extract light-invariant representations, enabling consistent and robust low-light image enhancement under diverse illumination conditions.
    
    \item We design two key modules to enhance illumination adaptability: LQM, which extracts and quantizes light factors to build light-invariant feature representations, and LAPM, which dynamically refines feature representations based on light-specific priors. These modular design enables stable and consistent feature representations across varying illumination conditions.

    \item We conduct extensive experiments in datasets, including LOL-v2-Real, LOL-v2-Synthetic, LSRW-Huawei, LIME, MEF, VV, and NPE, demonstrating that LightQANet consistently achieves state-of-the-art performance, as evidenced by the results shown in Fig.~\ref{fig:codebook} (c) and (d).
    
\end{itemize}

The remainder of this paper is organized as follows. Section \ref{related_works} reviews related works. Section \ref{methodology} presents a detailed model design. Section \ref{experiments} conducts the experiments and discusses the results. Finally, we conclude the paper in Section \ref{conclusion}.

\section{Related Works} \label{related_works}
This section reviews previous work related to low light image enhancement and recent advances in discrete codebook learning for image restoration tasks.

\subsection{Low-Light Image Enhancement}
Images captured in low-light environments typically suffer from poor quality, lacking essential visual details which can hinder comprehension and analysis. Initially, researchers tackled this problem through histogram equalization technologies \cite{AHE}, adjusting illumination and contrast by equalizing pixel intensity distributions. Various methods evolved from this approach, focusing on different enhancement aspects such as overall image perspective \cite{5773086}, cumulative distribution functions \cite{841534}, and the addition of penalty terms to refine the enhancement process \cite{4895264}.
Parallel to these developments, some researchers applied Retinex theory \cite{Land1971LightnessAR}, which decomposes an image into illumination and reflection components, allowing for targeted enhancements in both areas. Subsequent Retinex-based enhancements, such as SSR \cite{SSR}, improved both illumination and color accuracy significantly.
The advent of deep learning \cite{wen2021comprehensive} introduces more methods into the LLIE \cite{9743313}. LLNet \cite{LLNet} is the first to integrate stacked autoencoders for enhancing low-light images. This is followed by the introduction of multi-branch \cite{jin2025mb}\cite{liu2024dmfourllie} and multi-stage \cite{LRCR} networks, designed to tackle illumination recovery, noise suppression, and color refinement concurrently. SNR \cite{SNR} uses the PSNR distribution map to guide network feature learning and fusion. SMG \cite{SMG} incorporates image structural information to enhance the output image's quality.
Recent innovations have combined Retinex theory with deep learning to further refine enhancement techniques. URetinexNet \cite{URetinex} formulates the decomposition problem of Retinex as an implicit prior regularization model, and Retinexformer \cite{Retinexformer} uses illumination to guide the Transformer \cite{Transformer} in learning the global illumination information of the image. 
LLformer \cite{LLformer} proposes a new transformer and attention fusion block for LLIE.
GSAD \cite{GSAD}, JoRes \cite{wu2024jores} and LLDiffusion \cite{wang2025lldiffusion} leverage the diffusion model to perform LLIE.
QuadPrior~\cite{quadprior} improves low-light images by physical quadruple priors. CIDNet~\cite{CIDNet} proposed a new color space to overcome color bias and brightness artifacts in LLIE.

\subsection{Discrete Codebook Learning} 
Discrete codebook learning is first introduced in the context of Vector Quantized-Variational AutoenEoder (VQ-VAE) \cite{VQVAE}. Subsequently, VQ-GAN integrates this approach within the generative adversarial network framework, facilitating the generation of high-quality images \cite{VQGAN}. In low-level image processing tasks, codebook learning helps mitigate uncertainty during model training by transforming the operational space from raw images to a compact proxy space \cite{Codeformer}. To enhance feature matching, FeMaSR \cite{FeMaSR} introduces residual shortcut connections, RIDCP \cite{RIDCP} develops a controllable feature matching operation, and CodeFormer \cite{Codeformer} employs a Transformer-based prediction network for retrieving codebook indices. Additionally, LARSR \cite{LARSR} proposes a local autoregressive super-resolution framework utilizing the learned codebook. CodedBGT \cite{ye2024codedbgt} and CodeEnhance \cite{xu24} introduce the codebook to improve LLIE model performance. 
Different from CodedBGT \cite{ye2024codedbgt} and CodeEnhance \cite{xu24}, we propose the LQM to precisely extract light factors from image features. Additionally, we introduce the LAPM to dynamically enhance image representations through light-specific knowledge. Collectively, these modules significantly improve the representation of light information, thereby elevating the overall quality of the enhanced images.

\section{Methodology} \label{methodology}
This section provides a detailed introduction to the proposed method, which includes high-quality codebook learning, light-invariant feature learning, feature matching and image restoration, and training objectives.

\begin{figure*}[!t]
    \centering
    \includegraphics[width=0.95 \textwidth]{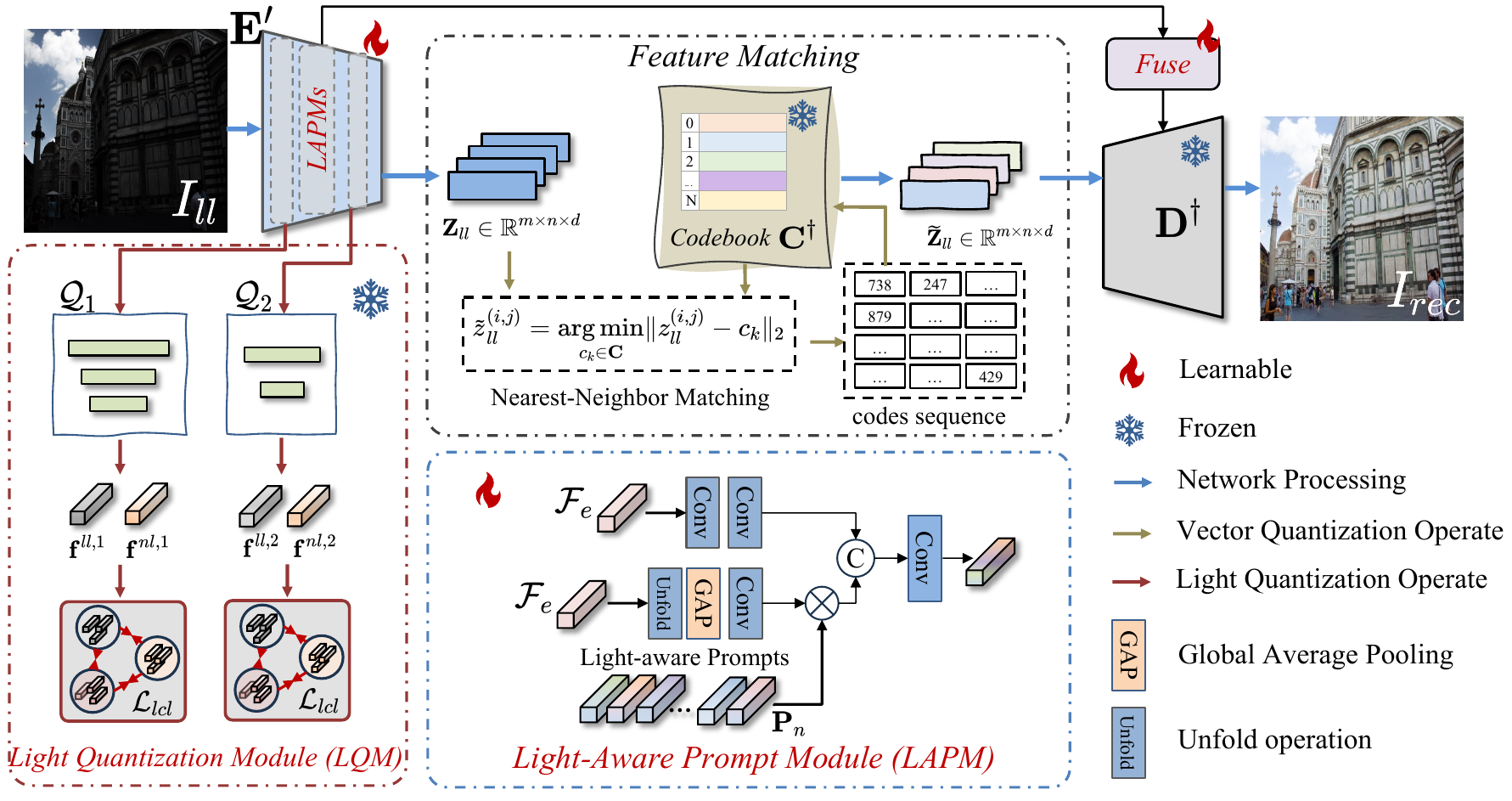}
    \caption{
    Overview of our proposed LightQANet framework. Our method leverages a pretrained codebook $\mathbf{C}^{\dag}$ and a decoder ${\rm D}^{\dag}$ as the foundations. In LightQANet, the Light Quantization Module (LQM) is utilized to extract light factors and promote the learning of light-invariant features by a novel light consistency loss ($\mathcal{L}_{lcl}$). Additionally, the Light-Aware Prompt Module (LAPM) is introduced to encode light illumination data for dynamically guiding the feature learning process. Finally, the feautre fusion via linear interpolation refines the reconstructed features.
}
\label{fig:framework}
\vspace{-4mm}
\end{figure*}

\subsection{Overview} \label{overview}
The proposed method consists of two stages: the first stage constructs a high-quality codebook using VQ-GAN trained on well-lit images to capture representative visual patterns. The second and more critical stage focuses on enhancing low-light images by extracting light-invariant features, ensuring stable representation and effective illumination correction across diverse lighting conditions.
Firstly, we leverage VQ-GAN to encode detailed features from high-quality images $I_h$ into a discrete set of codebook, which serve as a comprehensive reference for accurately reconstructing images. This stage can be formulated as follows:
\begin{equation}
\begin{aligned}
    &\mathbf{Z}_h = \mathrm{E}(I_h), \\ 
    &\mathbf{\widetilde{Z}}_{h} = \mathcal{M}(\mathbf{Z}_h, \mathbf{C}), \\
    &I'_h = \mathrm{D}(\mathbf{\widetilde{Z}}_{h}),
\end{aligned}
\end{equation}
where $\mathrm{E}(\cdot)$ and $\mathrm{D}(\cdot)$ denote encoder and decoder. $\mathcal{M}(\cdot, \cdot)$ is feature matching operation, where $\mathbf{C}$ represents learnable codebook of features. $\mathbf{Z}_h$ and $\mathbf{\widetilde{Z}}_{h}$ are latent features and quantized features. In the subsequent step, the $\mathbf{C}$ and $\mathrm{D}(\cdot)$ will be frozen to leverage the quantized features obtained from $\mathbf{C}$, followed by $\mathrm{D}(\cdot)$ reconstructing high-quality images. This ensures stability in the learning process and consistency in the output quality.

Next, as shown in Fig. \ref{fig:framework}, to improve feature extraction in low-light conditions, we develop light-invariant feature learning, where the LQM and LAPM are crucial for normalizing the impact of different lighting conditions on feature extraction, thereby stabilizing the model's performance across varied environments. 
Following this, we introduce feature matching and image reconstruction, examining how the algorithm aligns features under diverse illumination settings and reconstructs high-quality images from these aligned features. Low-light enhancement stage can be formulated as follows:
\begin{equation}
\begin{aligned}
    &\mathbf{Z}_{ll} = \mathrm{E}'(I_{ll}), \\
    &\mathbf{\widetilde{Z}}_{ll} = \mathcal{M}(\mathbf{Z}_{ll}, \mathbf{C}^{\dag}), \\
    &I_{rec} = \mathrm{D}^{\dag}(\mathbf{\widetilde{Z}}_{ll}, \mathbf{F}_{fuse}), \\
\end{aligned}
\end{equation}
where $I_{ll}$ and $I_{rec}$ denote low-light images and reconstructed images, respectively. $\mathbf{Z}_{ll}$ and $\mathbf{\widetilde{Z}}_{ll}$ are latent features and quantized features of $I_{ll}$. $\mathbf{F}_{fuse}$ represents output of the feature fusion in skip connection and is defined in Section \ref{FMIR}. 
${\rm E}'$ denotes the light-invariant feature learning.
$\mathbf{C}^{\dag}$ and $\mathrm{D}^{\dag}(\cdot)$ are codebook and decoder with frozen parameters, respectively.

\subsection{High-Quality Codebook Learning}
We first pre-train a VQ-GAN using high-quality images to learn a discrete codebook. 
This codebook serves as prior knowledge for enhancing low-light images. The corresponding decoder associated with the codebook is then utilized to reconstruct images.
Given a high-quality image $I_{h}$, we first employ the encoder of VQ-GAN to obtain a latent feature $\mathbf{Z}_h\in\mathbb{R}^{m \times n \times d}$. Then, by calculating the distance between each `pixel' $z_h^{(i,j)}$ of $\mathbf{Z}_h$ and the $c_k$ in the learnable codebook $\mathbf{C}=\{c_k \in \mathbb{R}^d\}_{k=0}^{N}$, we replace each $z_h^{(i,j)}$ with the nearest $c_k$ \cite{Codeformer}. After that, the quantized features $\mathbf{\widetilde{Z}}_{h} \in \mathbb{R}^{m\times n \times d}$ are obtained:
\begin{equation}\label{qua}
    \widetilde{z}^{(i,j)}_{h} =  \mathcal{M}(z^{(i,j)}_h, \mathbf{C}) = \mathop{\arg \min}\limits_{c_k \in \mathbf{C}} \Vert z^{(i,j)}_h - c_k\Vert_2 ,
\end{equation}
where $N=1024$ represents the size of the codebook, and $d=512$ denotes the channel number of both $\mathbf{Z}_h$ and $\mathbf{C}$. The dimensions $m$ and $n$ specify the sizes of $\mathbf{Z}_h$ and $\mathbf{\widetilde{Z}}_{h}$. The reconstructed image $I'_h$ is then generated by the decoder. The VQ-GAN is supervised using the loss function $\mathcal{L}_{vq}$ \cite{VQGAN}, which includes an $\mathbf{L}_1$ loss $\mathcal{L}_{mae}$, a codebook matching loss $\mathcal{L}_{cma}$, and an adversarial loss $\mathcal{L}_{adv}$:

\begin{equation}\label{vq_loss}
\begin{aligned}
  &\mathcal{L}_{vq} = \mathcal{L}_{mae} + \mathcal{L}_{cma} + \mathcal{L}_{adv},  \\ 
     &\mathcal{L}_{L1} = \Vert I_h - I'_h \Vert_1, \\ 
     &\mathcal{L}_{cma} = \sigma \Vert \mathbf{Z}_h - 
     \mathrm{sg}(\mathbf{\widetilde{Z}}_{h})\Vert_2^2 + \Vert \mathrm{sg}(\mathbf{Z}_h) - \mathbf{\widetilde{Z}}_{h} \Vert_2^2,  \\ 
     &\mathcal{L}_{adv} = \gamma \mathrm{log} \mathcal{D}(I_h) + \mathrm{log}(1 - \mathcal{D}(I'_h)),  
\end{aligned}
\end{equation}
where $\mathcal{D}(\cdot)$ is the discriminator. $\mathrm{sg}(\cdot)$ is the stop-gradient operator. $\sigma = 0.25$ denotes a weight trade-off parameter that governs the update rates of both the encoder and codebook \cite{Codeformer}. $\gamma$ is usually set to $0.1$ \cite{RIDCP}.

\begin{figure*}[!t]
    \centering
    \includegraphics[width=0.9 \textwidth]{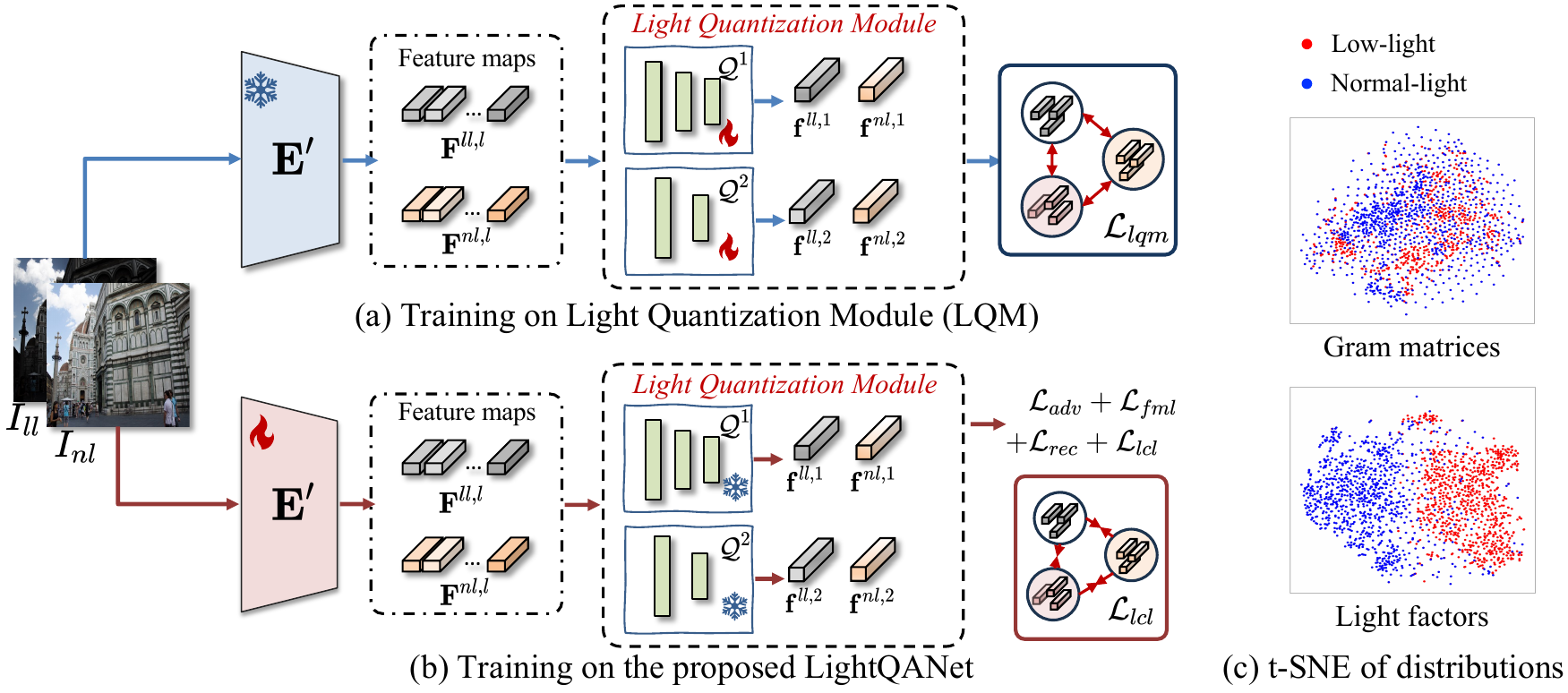}
    \caption{
    Overview of the proposed LQM. (a) and (b) illustrate the alternating optimization of the LQM and the LightQANet. 
    (c) shows the Gram matrices from low-light (LL) and normal-light (NL) images, which exhibit significant overlap. In contrast, the light factors are clearly distinguished based on lighting conditions, underscoring their effectiveness in accurately capturing and representing light-relevant information. 
    Note that the LQM is only used during the training phase and therefore does not impact the processing speed during the inference stage.
    }
\label{fig:LQM}
\vspace{-4mm}
\end{figure*}

\subsection{Light-Invariant Feature Learning} \label{LIFL}
The efficacy of our method depends on the quality of light-invariant feature learning.
To achieve this, we design two key modules: the LQM, which models illumination in a structured manner, and the LAPM, which adaptively guides feature learning based on illumination priors.

\textit{1) Light Quantization Module (LQM).}
To effectively extract light-invariant features for low-light image enhancement, we propose the LQM, motivated by the critical need to disentangle illumination information from detailed content representations.
Unlike conventional methods that operate directly on raw image features, we explicitly model illumination as a global style attribute, leveraging its inherent characteristics such as overall brightness, contrast, and color distribution, which are largely independent of fine-grained image details \cite{dumoulin2016learned}.
Drawing inspiration from the success of style-based approaches in nighttime domain adaptation \cite{gao2022cross} and low-light enhancement such as EnlightGAN \cite{EnlightGAN}, we adopt a Gram matrix \cite{gram} representation within LQM to capture and isolate illumination-related style features.
The Gram matrix effectively encodes global feature correlations, allowing LQM to abstract illumination information while filtering out irrelevant content-specific variations.
 The Gram matrix is defined as follows:
\begin{equation}
    \mathbf{G} = \mathbf{a}^{\rm T} \mathbf{a},
\end{equation}
where $\mathbf{G} \in \mathbb{R}^{c \times c}$, $c$ is the number of channels. $\mathbf{a}$ represents feature maps.

To equip the LQM with the ability to quantify illumination conditions, we formulate a supervised learning objective based on pairwise light factor distances.
Specifically, for a set of image pairs $\mathcal{P}$, LQM learns to construct a light factor space in which images captured under similar lighting conditions are mapped closer together, while those under different lighting conditions are pushed further apart.
The light factors $\mathbf{f}^{a,l}$ and $\mathbf{f}^{b,l}$ are computed by applying the LQM $Q(\cdot)$ to the Gram matrices $\mathbf{G}^{a,l}$ and $\mathbf{G}^{b,l}$ extracted from the $l$-th intermediate feature maps.
The training is supervised using the following loss:
\begin{equation} \label{lqm_loss}
    \begin{aligned}
        \mathcal{L}_{lqm} = \sum_{(a,b) \in \mathcal{P}} \Big\{ 
        (1 - \mathbf{1} (a, b)) \left[m - d(\mathbf{f}^{a,l}, \mathbf{f}^{b,l})\right]^2_+ \\
        + \mathbf{1} (a, b) \left[d(\mathbf{f}^{a,l}, \mathbf{f}^{b,l}) - m\right]^2_+\Big\},
    \end{aligned}
\end{equation}
where $d(\cdot)$ denotes the cosine similarity, $[\cdot]_+$ is the hinge function, $m$ is the margin, and $1(a,b)$ is an indicator function that returns 1 if $I^a$ and $I^b$ have the same lighting condition and 0 otherwise.
During this training stage, the encoder of the proposed is frozen and only the parameters of the LQM are updated, as shown in Fig.~\ref{fig:LQM} (a).
This learning process enables LQM to accurately quantify illumination differences and establish a light factor space is used to guide light-invariant feature learning in subsequent LightQANet training.

After LQM has acquired the ability to quantify illumination, we leverage its learned light factor space to guide the training of the proposed model.
Specifically, we introduce a light consistency loss $\mathcal{L}_{lcl}$ to minimize the discrepancy between the light factors of low-light and normal-light images.
As illustrated in Fig.~\ref{fig:LQM} (b), during this stage, the LQM is kept frozen while the encoder parameters are updated.
The light consistency loss is defined as:
\begin{equation} \label{lcl_loss}
    \mathcal{L}_{lcl}^l({\mathbf{f}^{a,l}}, {\mathbf{f}^{b,l}}) = \frac{1}{4d_l^2n_l^2} \sum_{i=1}^{d_l}({\mathbf{f}_i^{a,l}} - {\mathbf{f}_i^{b,l}})^2,
\end{equation}
where ${\mathbf{f}^{a,l}}$ and ${\mathbf{f}^{b,l}}$ are the light factors extracted by the frozen LQM, $d_l$ denotes the dimensionality of the light factors, and $n_l$ is the spatial size of the $l$-th feature map.
By minimizing $\mathcal{L}_{lcl}$, the encoder is encouraged to extract feature representations that are invariant to illumination variations, thus improving overall enhancement in various illumination scenarios.

\begin{algorithm}[!ht]
\caption{LightQANet Training Algorithm}
\begin{algorithmic}[1] 
\State \textbf{Input:} Paired training data $(I_{ll}, I_{nl})$: low-light and normal-light images
\State Randomly initialize the model parameters \( \theta \);
\For{each training pair ($I_{ll}$, $I_{nl}$)}
    \State $\mathbf{Z}_{ll} \leftarrow {\rm E}'(I_{ll})$;
    \State $\mathbf{\widetilde{Z}}_{ll} \leftarrow \mathcal{M}(\mathbf{Z}_{ll}, \mathbf{C}^{\dag})$;
    \State $I_{rec} \leftarrow \mathrm{D}^{\dag}(\mathbf{\widetilde{Z}}_{ll}, \mathbf{F}_{fuse})$;
    
    \State \textit{\# Update LQM and freeze LightQANet;}
    \State Optimize LQM using $\mathcal{L}_{lqm}$;
    \State \textit{\# Update LightQANet and freeze LQM;}
    \State Optimize LightQANet using combined loss: 
    $\mathcal{L}_{adv} + \mathcal{L}_{fml} + \mathcal{L}_{rec} + \mathcal{L}_{lcl}$;
\EndFor
\State \textbf{Return} Enhanced image $I_{rec}$
\end{algorithmic}
\label{algorithm1}
\end{algorithm}

Algorithm 1 and Fig.~\ref{fig:LQM} (a) and (b) illustrate the alternating optimization process between the LQM and the LightQANet network. This optimization strategy gradually reduces the discrepancy in illumination conditions between low-light and normal-light images, ultimately promoting the extraction of light-invariant features within the LightQANet framework.
To demonstrate the effectiveness of LQM, we analyze the Gram matrices computed from intermediate feature maps and their corresponding light factors produced by LQM. As shown in Fig.~\ref{fig:LQM} (c), the LQM effectively isolates illumination information, clearly separating it from other content-related features. These results indicate that the extracted light factors successfully encode illumination-specific attributes, as intended.

\textit{2) Light-Aware Prompt Module (LAPM).}
While LQM effectively captures structured illumination characteristics, it faces limitations in representing complex and spatially varying lighting patterns typically observed in real-world scenarios. To overcome this, LAPM dynamically adapts feature representations by aggregating illumination information from local spatial regions. Specifically, LAPM computes prompt weights based on local features rather than relying solely on a global illumination descriptor. This enables each region in the image to contribute effectively to the dynamic prompt composition, thus capturing fine-grained variations in illumination and providing more adaptive modulation of the final feature representation.

Fig. \ref{fig:framework} shows that the prompt component $\mathbf{P}_n$, consisting of five learnable vectors, embeds light information from $n$ levels. 
These prompt vectors are not only responsible for modeling discrete brightness states but are also trained to encode transitional relationships between different brightness levels and to capture global illumination properties.
To generate the light-aware prompts $\mathbf{P}$, we compute attention-based weights from local features and then apply these weights to $\mathbf{P}_n$. 
The weights serve as region-wise “soft assignments,” guiding each prompt to specialize in the luminance ranges where it is most effective (e.g., extremely dark vs. mid-level brightness). Summing these weighted prompts yields a prompt-guided feature modulation that faithfully reflects the overall illumination distribution, from darkest shadows to brightest highlights.
Specifically, we first divide the image features into $n$ patches. Average pooling is applied to these patches to extract local features, which are then processed by a channel-shrink layer to ensure their dimensions align with $\mathbf{P}_n$. After dimension alignment, a softmax function denoted as ${\rm F_{s}}$, is employed to compute the weights $\omega_n \in \mathbb{R}^C$. The weights interact with the $\mathbf{P}_n$ to generate $\mathbf{P}$, which are further processed by a convolution layer with a $3 \times 3$ kernel size. These operations can be collectively formulated as follows:
\begin{equation}
\begin{aligned}
\mathbf{P} &= {\rm LAPM}(\mathbf{F}_l, \mathbf{P}_n)
={\rm F}_{3} (\sum_{n=1}^N \omega_n \mathbf{P}_n), \\
\omega_n &= {\rm F}_{s} ({\rm F}_{1}({\rm F}_{A} (\mathbf{F}_l))),
\end{aligned}
\end{equation}
where $\mathbf{F}_l = {\rm UF}(\mathbf{F}_e)$ represents local features. ${\rm UF}(\cdot)$ is a unfold operation. $\mathbf{F}_e$ denotes the intermediate features of encoder. ${\rm F}_A(\cdot)$ is a average pooling operation. ${\rm F}_1(\cdot)$ is the channel-shrink layer performed by a $1 \times 1$ convolution layer. 
Finally, the light-aware prompts $\mathbf{P}$ are integrated channel-wise with the intermediate features of the encoder. These combined features are then processed by a ResNet block \cite{resnet}, enhancing the overall feature representation.

\begin{figure}[!t]
    \centering
    \includegraphics[width=0.48 \textwidth]{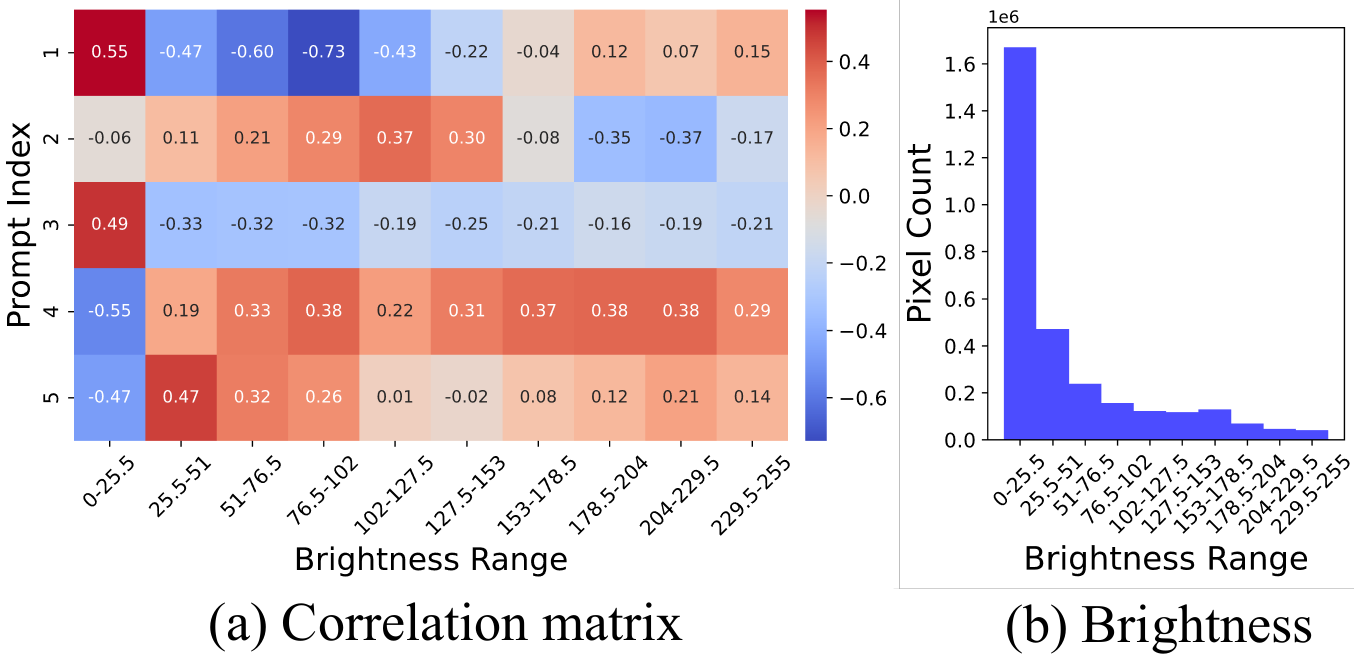}
    \caption{
    Prompt-Brightness correlation analysis.
    (a) Correlation heatmap between prompt weights (Prompt Index 1–5) and brightness levels. Positive (red) and negative (blue) values indicate correlation strength. Different prompts show distinct sensitivities, with some strongly responding to extremely dark regions [0, 25.5) and others to brighter or transitional regions, highlighting their complementary roles. (b) Brightness distribution histogram of the MEF dataset \cite{MEF}, showing a predominance of darker pixels.
    }
    \label{fig:prompt_correlation}
    \vspace{-4mm}
\end{figure}

Fig.~\ref{fig:prompt_correlation} (a) shows that each prompt vector responds distinctly to different brightness ranges. For example, prompts 1 and 3 are highly sensitive (correlations of 0.55 and 0.49) to extremely dark pixels ([0, 25.5)). Prompt 5 primarily handles low-light pixels [25.5, 51), showing a correlation of 0.47. Prompt 4 mainly focuses on mid-to-high brightness levels (above 76.5), effectively capturing general illumination patterns. Prompt 2 complements these prompts by responding moderately (correlation of 0.37) to intermediate brightness levels [102, 127.5), and mild negative correlations in extremely dark [0, 25.5) and bright regions (above 178.5).
Furthermore, our prompt allocation strategy is informed by the illumination conditions, as illustrated by Fig.~\ref{fig:prompt_correlation} (b). Since the majority of pixels in low-light images fall into the darkest brightness interval ([0,25.5)), we assign two dedicated prompts (prompts 1 and 3) to this interval. 
Overall, our multi-prompt vector design and data-driven allocation strategy ensure each prompt effectively captures illumination-specific information, especially for diverse brightness conditions.

\subsection{Feature Matching and Image Reconstruction}\label{FMIR}
We perform feature matching through nearest-neighbor lookup in the codebook to obtain high-quality features, as shown in Fig. \ref{fig:framework}.
Subsequently, these high-quality features are transmitted to a decoder that incorporates skip feature fusion modules, enabling the reconstruction of enhanced images. 
Based on Eq. \ref{qua}, the feature matching $\mathcal{M}(\cdot,\cdot)$ in low-light enhancement task can be formulated as follows:
\begin{equation}\label{qua_ll}
    \widetilde{z}^{(i,j)}_{ll} =  \mathcal{M}(z^{(i,j)}_{ll}, \mathbf{C}^{\dag}) = \mathop{\arg \min}\limits_{c_k \in \mathbf{C}^{\dag}} \Vert z^{(i,j)}_{ll} - c_k\Vert_2 ,
\end{equation}
where $\mathbf{Z}_{ll} = \{z_{ll}^{(i,j)} \in \mathbb{R}^{d} \} _{i=0, j=0}^{m,n}$ denotes image features extracted by light-invariant feature learning, $\mathbf{\widetilde{Z}}_{ll} = \{\widetilde{z}_{ll}^{(i,j)} \in \mathbb{R}^{d} \} _{i=0, j=0}^{m,n}$ represents quantized features.

To improve the quality of reconstructed images, we introduce a feature fusion via linear interpolation technique that effectively merges low-level features $\mathbf{F}_e$ from the encoder with features $\mathbf{F}_d$ from the decoder. This integration not only preserves critical texture information but also compensates for potential detail loss during image processing. 
Initially, features $\mathbf{F}_e$ and $\mathbf{F}_d$ are combined in channel-wise and subsequently computes affine transformation parameters $\bm \alpha$ and $\bm \beta$. 
These parameters are designed to reduce the impact of noise and enhance texture representation in the reconstructed images. 
This feature fusion can be formulated as follows:

\begin{equation}
\begin{aligned}
    \mathbf{F}_{fuse} &= \bm{\alpha} \odot \mathbf{F}_d  + \bm{\beta}, \\
    \bm{\alpha}, \bm{\beta} &= \mathcal{C}([\mathbf{F}_d,\mathbf{F}_e]),
\end{aligned}
\end{equation}
where $\mathcal{C}(\cdot)$ denotes convolution operation, and $\odot$ is element-wise multiplication. Finally, we employ the LAPM to further refine the features. Importantly, the parameters within the decoder blocks remain frozen during this process.

\begin{table*}[]
\centering
\small
\caption{Quantitative results on LOL-v2 \cite{LOLv2} and LSRW-Huawei \cite{LSRW}. $\uparrow$ indicates the higher the better. $\downarrow$ indicates the lower the better. $\textbf{Bold}$: Best result; {\ul{Underline}}: Second best result; $-$: Unavailable data. P and F denotes parameters and FLOPs.}
\resizebox{1 \textwidth}{!}{
\begin{tabular}{l|cccc|cccc|cccc|cc}
\toprule
 & \multicolumn{4}{c|}{LOL-V2-Real} & \multicolumn{4}{c|}{LOL-V2-Synthetic} & \multicolumn{4}{c|}{LSRW-Huawei} & \multicolumn{2}{c}{Complexity} \\
 \multirow{-2}{*}{Methods}       & PSNR $\uparrow$ & SSIM $\uparrow$ & LPIPS $\downarrow$ & NIQE $\downarrow$ & PSNR $\uparrow$ & SSIM $\uparrow$ & LPIPS $\downarrow$ & NIQE $\downarrow$ & PSNR $\uparrow$ & SSIM $\uparrow$ & LPIPS $\downarrow$ & NIQE $\downarrow$ & Param (M) & FLOPs (G) \\ \midrule  
LIME                      & 16.97          & 0.4598          & 0.3415          & {\color[HTML]{000000} 8.4899}          & 17.50          & 0.7718          & 0.1748          & {\color[HTML]{000000} 3.4063}          & 18.46          & 0.4450          & 0.3922          & {\color[HTML]{000000} 3.3879}          & -              & -             \\
JED                       & 17.29          & 0.7266          & 0.2760          & {\color[HTML]{000000} 4.3703}          & 16.89          & 0.7299          & 0.2370          & {\color[HTML]{000000} 3.5284}          & 15.11          & 0.5379          & 0.4327          & {\color[HTML]{000000} 3.1521}          & -              & -             \\
RetinexNet                & 16.10          & 0.4006          & 0.4215          & {\color[HTML]{000000} 9.2661}          & 17.14          & 0.7615          & 0.2185          & {\color[HTML]{000000} 4.3433}          & 16.82          & 0.3951          & 0.4566          & {\color[HTML]{000000} 3.4942}          & 0.84           & 587.47        \\
KinD                      & 16.75          & 0.6456          & 0.4118          & {\color[HTML]{000000} 4.6253}          & 17.51          & 0.7694          & 0.2093          & {\color[HTML]{000000} 3.3295}          & 17.19          & 0.4625          & 0.4318          & {\color[HTML]{000000} 2.9682}          & 8.02           & 34.99         \\
EnlightGAN                & 17.94          & 0.6755          & 0.3197          & {\color[HTML]{000000} 4.8755}          & 16.59          & 0.7780          & 0.2179          & {\color[HTML]{000000} 3.0998}          & 17.46          & 0.4982          & 0.3780          & {\color[HTML]{000000} 3.0650}          & 114.35         & 61.01         \\
Zero                      & 18.06          & 0.5736          & 0.2980          & {\color[HTML]{000000} 7.7571}          & 17.76          & 0.8163          & 0.1382          & {\color[HTML]{000000} 3.0464}          & 16.40          & 0.4761          & 0.3763          & {\color[HTML]{000000} 3.0477}          & 0.075          & 4.83          \\
SNR                       & 21.48          & 0.8489          & 0.1996          & {\color[HTML]{000000} 3.6383}          & 22.88          & 0.8962          & 0.1124          & {\color[HTML]{000000} 3.5854}          & 20.67          & 0.6246          & 0.4879          & {\color[HTML]{000000} 3.4008}          & 39.12          & 26.35         \\
SMG                       & 24.03          & 0.8178          & 0.2283          & {\color[HTML]{000000} 5.7291}          & 25.62          & 0.9188          & 0.2915          & {\color[HTML]{000000} 5.9165}          & 20.66          & 0.5589          & 0.4449          & {\color[HTML]{000000} 6.9247}          & 0.33           & 20.81         \\
PairLIE                   & 19.88          & 0.7777          & 0.2834          & {\color[HTML]{000000} 3.6192}          & 19.07          & 0.7965          & 0.2183          & {\color[HTML]{000000} 3.9121}          & 18.99          & 0.5632          & 0.3711          & {\color[HTML]{000000} 3.0790}          & 1.61           & 15.57         \\
RetinexFormer             & 22.79          & 0.8397          & 0.2270          & {\color[HTML]{000000} 3.3869}          & 25.67          & 0.9296          & 0.0775          & {\color[HTML]{000000} 2.8861}          & 20.81          & 0.6303          & 0.4124          & {\color[HTML]{000000} 2.8866}          & 30.35          & 137.37        \\
CodeEnhance               & 23.32          & 0.8310          & 0.2184          & {\color[HTML]{000000} 3.2115}          & 24.65          & 0.9163          & 0.0648          & {\color[HTML]{000000} 3.2019}          & 21.14          & 0.6076          & \textbf{0.2840} & {\color[HTML]{000000} 2.6424}          & 49.07          & 225.86        \\
DMFourLLIE                & 22.64          & 0.8589          & {\ul 0.1488}    & {\color[HTML]{000000} \textbf{2.9389}} & {\ul 25.83}    & 0.9314          & 0.0562          & {\color[HTML]{000000} 2.8892}          & {\ul 21.47}    & {\ul 0.6331}    & 0.3998          & {\color[HTML]{000000} 3.0153}          & 0.41           & 1.56          \\
QuadPrior                 & 20.58          & 0.8036      f    & 0.2410          & {\color[HTML]{000000} 5.8903}          & 16.11          & 0.7646          & 0.2187          & {\color[HTML]{000000} 4.6653}          & 18.30          & 0.6013          & 0.4070          & {\color[HTML]{000000} 3.7033}          & 1252.75        & 1103.20       \\
CIDNet                    & {\ul 24.11}    & {\ul 0.8675}    & 0.1678          & {\color[HTML]{000000} 3.4159}          & 25.13          & {\ul 0.9387}    & {\ul 0.0536}    & {\color[HTML]{000000} \textbf{2.8128}} & 20.86          & 0.6202          & 0.3740          & {\color[HTML]{000000} {\ul 2.6131}}    & 1.88           & 7.57          \\
Ours                      & \textbf{28.51} & \textbf{0.8974} & \textbf{0.1039} & {\color[HTML]{000000} {\ul 3.1926}}    & \textbf{26.15} & \textbf{0.9388} & \textbf{0.0457} & {\color[HTML]{000000} {\ul 2.8636}}    & \textbf{21.68} & \textbf{0.7179} & {\ul 0.2885}    & {\color[HTML]{000000} \textbf{2.5784}} & 18.85          & 164.20       \\  \bottomrule 
\end{tabular}}
\label{table:compare}
\end{table*}

\begin{table}[!t]
\centering
\caption{Comparisons on LIME \cite{LIME}, MEF \cite{MEF}, NPE \cite{NPE}, and VV \ref{vv} in terms of NIQE, where the lower the better. Note that the results "Null" are due to the corresponding methods lacking code.}
\begin{tabular}{l|cccc}
\toprule
Methods       & LIME          & MEF           & NPE           & VV            \\
\midrule
KinD$_{2019}$          & 6.71          & 3.17          & \underline{3.28}          & 2.32          \\
EnlightGAN$_{2019}$    & \underline{3.59}          & \underline{3.11}          & 4.36          & 3.18          \\
Zero$_{2020}$          & 3.79          & 3.31          & 3.48          & 2.75          \\
SNR$_{2022}$           & 4.88          & 3.47          & 4.19          & 7.55          \\
PairLIE$_{2023}$       & 4.31          & 3.92          & 3.68          & 3.16          \\
RetinexFormer$_{2023}$ & 3.70          & 3.14          & 3.58          & \underline{1.95} \\
SMG$_{2023}$           & 6.47          & 6.18          & 5.89          & 5.46          \\
CodedBGT$_{2024}$      & 4.20          & 3.85          & 3.52          & Null          \\
QuadPrior$_{2024}$     & 4.58          & 4.36          & 3.65          & 3.44          \\
CIDNet$_{2025}$        & 3.85          & 3.46          & 3.82          & 3.24          \\ \midrule
Ours          & \textbf{3.54} & \textbf{2.88} & \textbf{3.26} & \textbf{1.94} \\ \bottomrule
\end{tabular}
\label{table:compare_unpaired}
\end{table}

\subsection{Training Objectives}
Finally, we outline the training objectives that guide the overall learning process. Specifically, the LQM is optimized with a dedicated contrastive loss ($\mathcal{L}_{lqm}$), which has defined in Eq. \ref{lqm_loss} in Section \ref{LIFL}. The main enhancement model is trained with a combination of Adversarial Loss $\mathcal{L}_{adv}$, Feature Matching Loss $\mathcal{L}_{fml}$, Light Consistency Loss $\mathcal{L}_{lcl}$, and Reconstruction Loss $\mathcal{L}_{rec}$ to ensure high-quality restoration under varying illumination conditions, which are defined as follows:

\begin{equation}
    \mathcal{L}_{total} = \mathcal{L}_{adv} + \mathcal{L}_{fml}+ \mathcal{L}_{rec} + \lambda \mathcal{L}_{lcl},
\end{equation}
where $\mathcal{L}_{adv}$ is defined in Eq. \ref{vq_loss} of Section \ref{overview}. $\lambda=0.5$ is a weight of $\mathcal{L}_{lcl}$. $\mathcal{L}_{lcl}$ is used to minimize the discrepancy between
the light factors of low-light and normal-light images, which has defined in Eq. \ref{lcl_loss} of Section \ref{LIFL}.

\textit{1) Feature Matching Loss.} 
This loss function is specifically designed to optimize the encoder by facilitating its learning of the mapping between low-light images and high-quality priors. By minimizing this loss, the encoder can enhance the proposed method ability to accurately translate low-light conditions into visually appealing outputs, aligning with predefined high-quality standards. The loss is formulated as follows:
\begin{align}
    \mathcal{L}_{fml} &= \sigma \Vert \mathbf{{Z}}_{ll} - 
    \mathrm{sg}(\mathbf{\widetilde{Z}}_{h})\Vert_2^2 + \Vert \phi(\mathbf{{Z}}_{ll}) - \phi(\mathrm{sg}(\mathbf{\widetilde{Z}}_{h})) \Vert_2^2 \notag
\end{align}
where $\phi(\cdot)$ is used to calculate the gram matrix of features. $\mathbf{{Z}}_{ll}$ and $\mathbf{\widetilde{Z}}_{h}$ represent the latent features of low-light images and quantized features of high-quality images, respectively.

\textit{2) Reconstruction Loss.} 
This loss function combines $\mathbf{L}_1$ loss and perceptual loss to ensure that enhanced images have a complete structure and impressive visual appeal. The $\mathbf{L}_1$ loss minimizes pixel-level discrepancies for high fidelity, while perceptual loss aligns images to human visual perception, enhancing both structural accuracy and aesthetic quality. The loss is defined as follows:
\begin{equation}
    \mathcal{L}_{rec} = \Vert I_{nl} - I_{rec} \Vert_1 + \Vert \psi(I_{nl}) - \psi(I_{rec}) \Vert_2^2,
\end{equation}
where $I_{nl}$ and $I_{rec}$ represent normal-light images and reconstructed images, respectively. $\psi(\cdot)$ indicates the Learned Perceptual Image Patch Similarity (LPIPS) function \cite{lpips}.

\section{Experiments} \label{experiments}
This section presents experimental results to evaluate the effectiveness of the proposed method through quantitative comparisons, qualitative analysis, and ablation studies.

\subsection{Implementation Details}
For VQ-GAN and the proposed LightQANet training, the input pairs are randomly cropped to patches of size 256 $\times$ 256. We use ADAM optimizer with $\beta_1 = 0.9$, $\beta_2 = 0.999$ and $\varepsilon=10^{-8}$. The learning rate is set to $10^{-4}$. The VQ-GAN is pre-trained on the DIV2K \cite{DIV2K} and Flickr2K \cite{Flickr2K} with 350K iterations. Our LightQANet is trained with 50K iterations. The hyper-parameter $m$ is set to 0.1. All experiments were conducted in PyTorch on an NVIDIA A6000.

\begin{figure*}[!t]
    \centering
    \includegraphics[width=0.95 \textwidth]{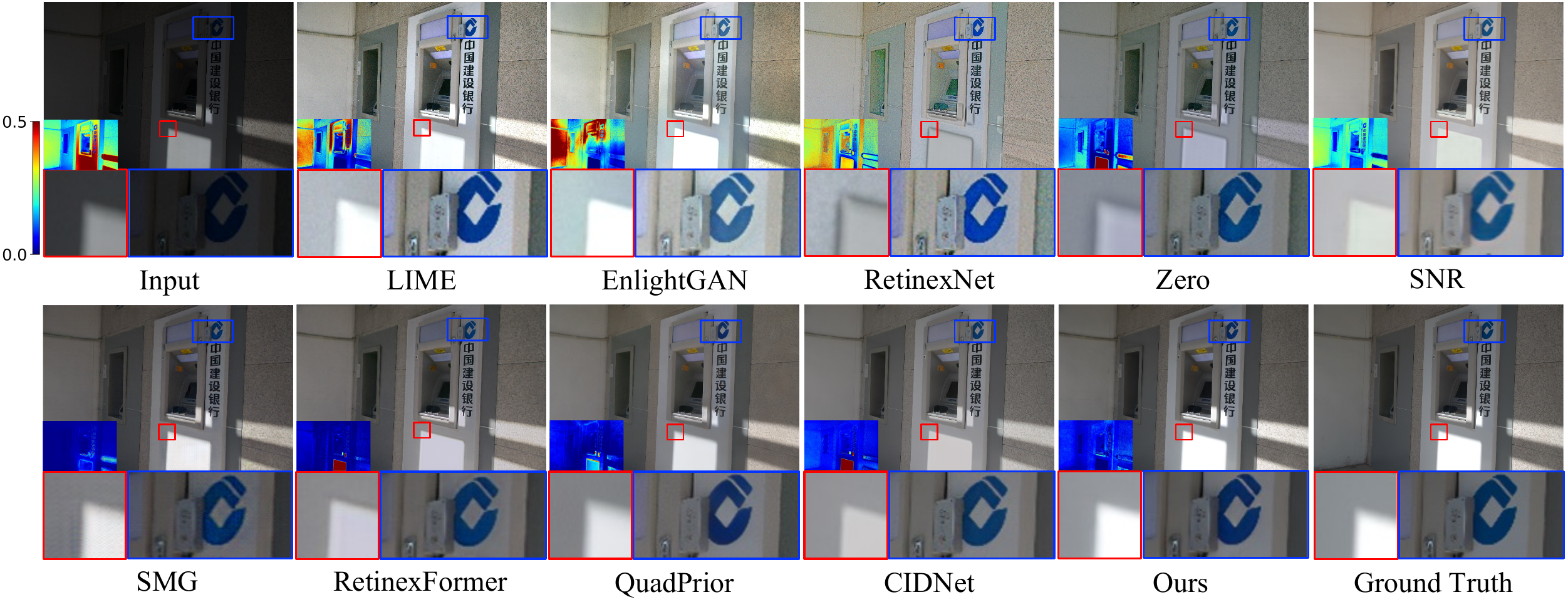}
    \vspace{-2mm}
    \caption{Visual comparison on the LOL-v2-Real \cite{LOLv2}, accompanied by image error maps calculated by $\mathrm{L2}$ loss. The proposed method produces a more natural illumination transition across shadow boundaries (e.g., wall region), accurate colors, and well-preserved details.}
    \vspace{-2mm}
    \label{fig:compare_LOL_real}
\end{figure*}
\begin{figure*}[!t]
    \centering
    \includegraphics[width=0.95 \textwidth]{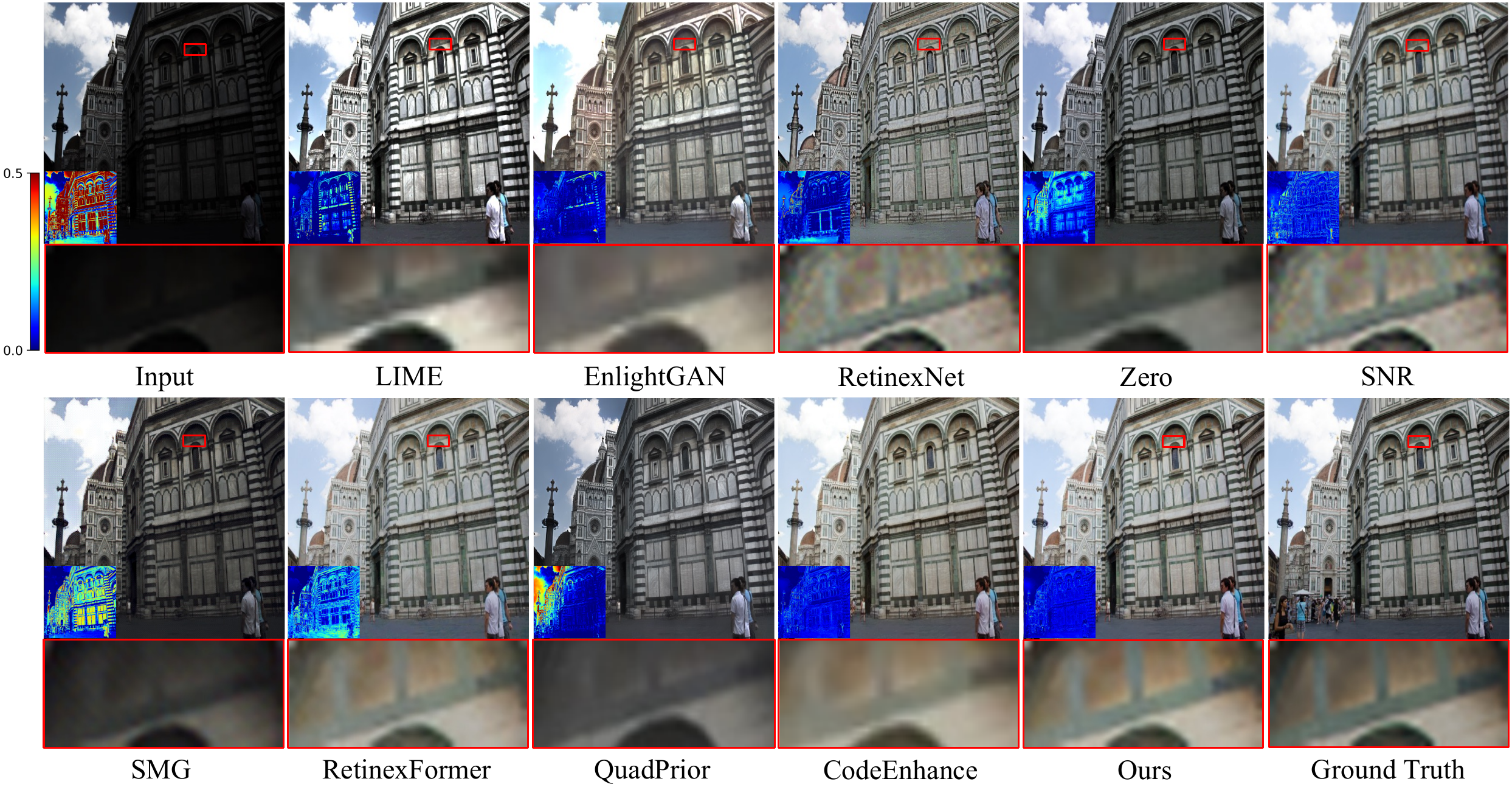}
    \caption{Visual comparison on the LOL-v2-Synthetic \cite{LOLv2}, accompanied by image error maps. Compared to competing methods, our approach demonstrates superior overall brightness and structural clarity. LIME, EnlightGAN, SMG, QuadPrior, and CodeEnhance display noticeable color deviations.}
    \vspace{-2mm}
    \label{fig:compare_LOL_syn}
\end{figure*}

\begin{figure*}[!t]
    \centering
    \includegraphics[width=0.95 \textwidth]{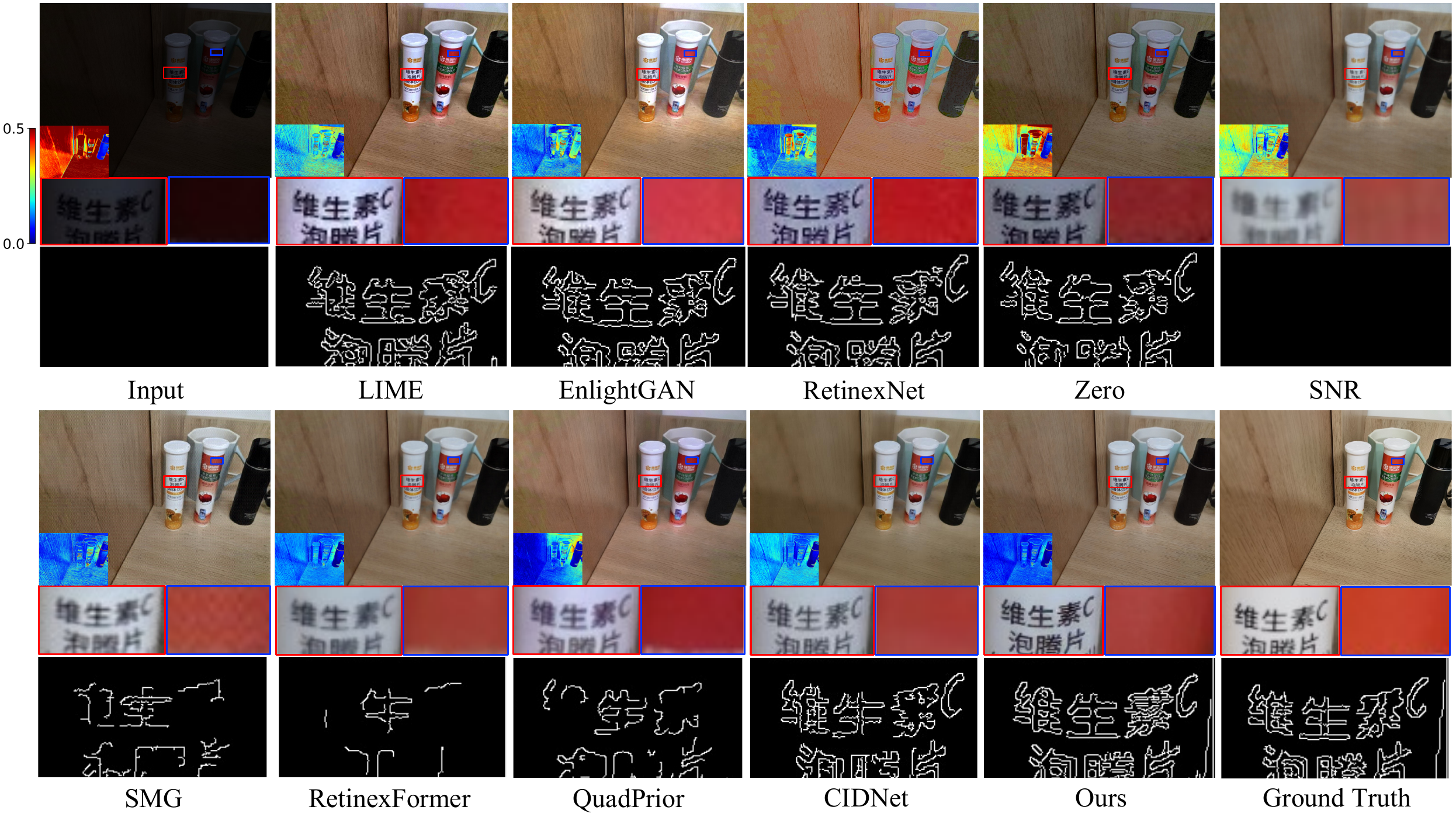}
    \caption{Visual comparison on the LSRW Huawei \cite{LSRW}, accompanied by image error maps and Canny edge maps of text regions. As we can see that the proposed method restores both texture and color details effectively. }
    \label{fig:compare_Huawei}
\end{figure*}
\vspace{-5mm}
\begin{figure*}[!t]
    \centering
    \includegraphics[width=0.9 \textwidth]{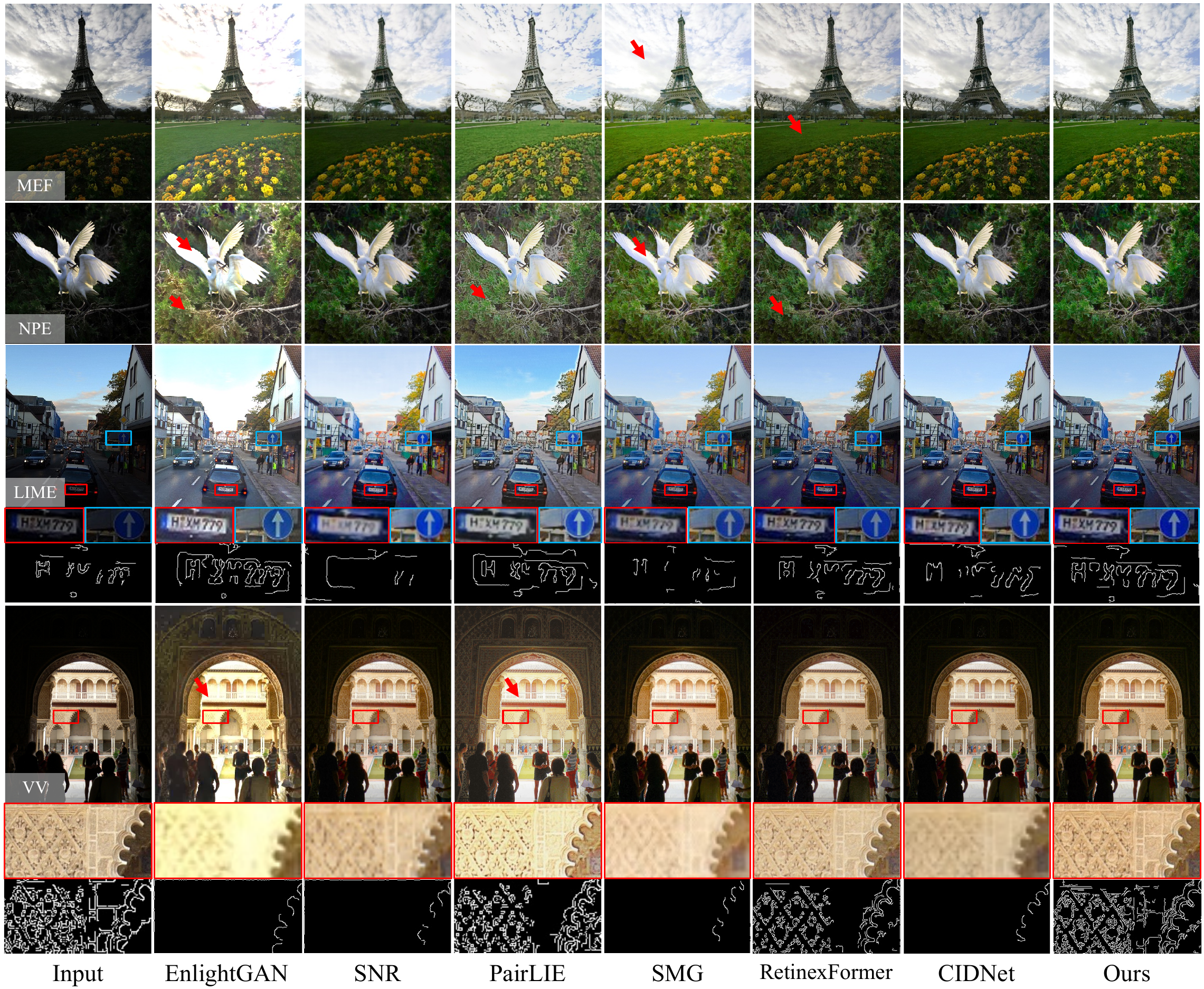}
    \caption{Visual comparison on the LIME \cite{PairLIE}, MEF \cite{MEF}, VV \ref{vv}, and NPE \cite{NPE} dataset, accompanied by Canny edge maps. The results demonstrate that our method effectively enhances images across different lighting conditions, achieving a trade-off between texture preservation and illumination enhancement.}
\label{fig:compare_unpaired}
\end{figure*}

\subsection{Datasets and Evaluation Metrics}
\textit{1) Low-light Datasets.} We evaluate methods using the LOL-v2 \cite{LOLv2} and LSRW-Huawei \cite{LSRW}. And also evaluate methods in cross datasets: LIME \cite{LIME}, MEF \cite{MEF}, VV \footnote{https://sites.google.com/site/vonikakis/datasets\label{vv}}, and NPE \cite{NPE}
The LOL-v2-Real one contains 689 train images and 100 test images. The LOL-v2-Synthetic one includes 900 train images and 100 test images. 
The LSRW-Huawei contains 2,450 train images and 30 test images. 
The LIME, MEF, VV, and, NPE include 10, 17, 24, and 8 low-light images, respectively.

\textit{2) Evaluation Metrics.}
We assess the quality of the enhanced images using the most common metrics: Peak Signal-to-Noise Ratio (PSNR), Structural Similarity Index Measure (SSIM) \cite{SSIM}, Natural Image Quality Evaluator (NIQE) \cite{NIQE}, and LPIPS \cite{lpips}. Unlike PSNR and SSIM, which primarily focus on lowe-level similiaity, LPIPS accounts for the human visual system's perception of similarity, offering a more accurate reflection of how images are perceived by viewers.

\subsection{Comparison with State-of-the-Art Methods}
We assess the performance of our LightQANet by conducting comparisons with numerous leading LLIE techniques. These include LIME \cite{LIME}, JED \cite{JED}, RetinexNet \cite{RetinexNet}, KinD \cite{KinD}, EnlightGAN \cite{EnlightGAN}, Zero \cite{Zero}, SNR \cite{SNR}, PairLIE \cite{PairLIE}, RetinexFormer \cite{Retinexformer}, SMG \cite{SMG}, CodedBGT \cite{ye2024codedbgt}, DMFourLLIE \cite{liu2024dmfourllie}, CodeEnhance~\cite{xu24}, QuadPrior~\cite{quadprior}, and CIDNet~\cite{CIDNet}. 

\textit{1) Same-Domain Evaluation.} 
We conduct detailed visual comparisons on LOL-v2-Real (see Fig.~\ref{fig:compare_LOL_real}), LOL-v2-Synthetic (see Fig.~\ref{fig:compare_LOL_syn}), and LSRW-Huawei (see Fig.~\ref{fig:compare_Huawei}), focusing on four key aspects, with quantitative results summarized in Table\ref{table:compare}.
\textit{Illumination Consistency:} Competing methods such as Zero, CIDNet, and RetinexFormer exhibit abrupt brightness transitions, while SMG, QuadPrior, and Zero under-enhance dark regions. Our method achieves smooth, natural illumination transitions, which is reflected in the highest SSIM scores (0.8974, 0.9388, 0.7179) across all datasets.
\textit{Color Fidelity:} LIME, RetinexNet, and EnlightGAN introduce strong color distortions; Zero, SMG, and QuadPrior show biases in wall and text regions. LightQANet restores accurate hues, aided by illumination quantization and adaptive modulation, yielding the lowest LPIPS values (0.1039, 0.0457, 0.2885).
\textit{Texture Preservation:} SNR, SMG, QuadPrior, and CIDNet fail to recover fine details, producing smoothed textures. Our method preserves sharp contours and structural details, consistent with the top PSNR results (28.51, 26.15, 21.68).
\textit{Artifact Suppression:} RetinexNet, RetinexFormer, and SNR introduce noise or halos; CodeEnhance causes over-smoothing. In contrast, LightQANet suppresses artifacts effectively, balancing enhancement and detail.

Across the three datasets, the proposed method consistently shows superior performance in handling illumination transitions, restoring accurate colors, preserving fine textures, and minimizing enhancement artifacts. These results further validate the effectiveness of LightQANet in delivering robust and perceptually pleasing low-light enhancement.

\textit{2) Cross-Domain Evaluation.} 
To assess the robustness and generalization ability of our network under domain shift, we evaluate the model on unpaired low-light images from datasets that differ from the training domain, including MEF \cite{MEF}, NPE \cite{NPE}, LIME \cite{LIME}, and VV \ref{vv}. As illustrated in Fig. \ref{fig:compare_unpaired} and Table \ref{table:compare_unpaired}, the proposed method consistently achieves superior visual quality across diverse scenes and lighting conditions.
Specifically, in the MEF dataset, our method restores natural brightness and preserves fine textures in both the grass and flower regions, while other methods tend to overexpose the sky or blur the foreground textures (see red arrows). In the NPE scene, we accurately enhance the dove's feathers and surrounding foliage without color distortion or effects, unlike methods such as EnlightGAN and PairLIE which introduce unatural brightness or lose edge sharpness.
In the LIME dataset, the proposed method clearly reveals the license plate characters and traffic sign symbols (highlighted in the red box), which are over-smoothed (SNR, SMG) in other methods. For the VV dataset, our method demonstrates superior detail preservation in both global structure and fine textures. Notably, the intricate wall carvings (see bottom row) are sharp and realistic in our result, whereas others either blur the details (SNR, SMG, and CIDNet) or color-destoration the region (EnlightGAN, PairLIE).

These improvements can be attributed to the combination of structured modeling of illumination and high-quality prompt tailored for low-light conditions. Together, they allow the model to adaptively enhance illumination while maintaining structural fidelity and natural appearance, even when applied to previously unseen domains.

\textit{4) Complexity Analysis.}
Table~\ref{table:compare} details model complexity (Params, FLOPs) at 256×256 resolution. Simple CNNs like Zero (0.075M params, 4.83G FLOPs) and DMFourLLIE offer low complexity but compromise enhancement quality. Conversely, QuadPrior's use of a pretrained diffusion backbone leads to high complexity (1252.75M params, 1103.2G FLOPs). Our LightQANet achieves a strong balance with 18.9M parameters and 164.2G FLOPs, representing a significant 61\% reduction in parameters and 27\% in FLOPs compared to the codebook-based CodeEnhance, thus offering superior quality at a moderate computational cost.

\begin{figure*}[!t]
    \centering
    \includegraphics[width=0.9 \textwidth]{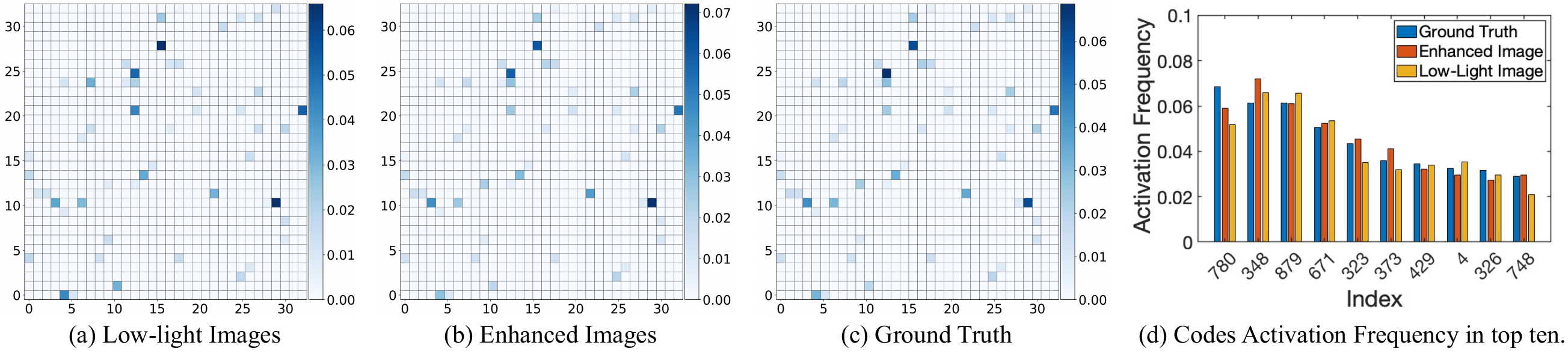}
    \caption{
    Comparison of code activation frequency. The codebook includes 1024 quantized features, we reshape these feature indexes into $32 \times 32$. (a) denotes the code activate frequency of low-light images under our method.
    (b) and (c) represent code activation frequency of enhanced images and ground truth under baseline. (d) is a comparison of the top ten codes' activation frequency. 
    The analysis shown in (a), (c), and (d) highlights the capability of the proposed method to effectively extract features from low-light conditions, achieving activation frequencies closely aligned with those of the ground truth. Additionally, (b), (c), (d) demonstrate the high quality of our results, exhibiting similar activation patterns to the ground truth within the baseline comparisons. This coherence across the figures substantiates the effectiveness of our enhancement approach in maintaining the integrity of image features under varied lighting conditions.
    }
\label{fig:Codebook_index_count}
\end{figure*}

\begin{figure*}[!t]
    \centering
    \includegraphics[width=0.9 \textwidth]{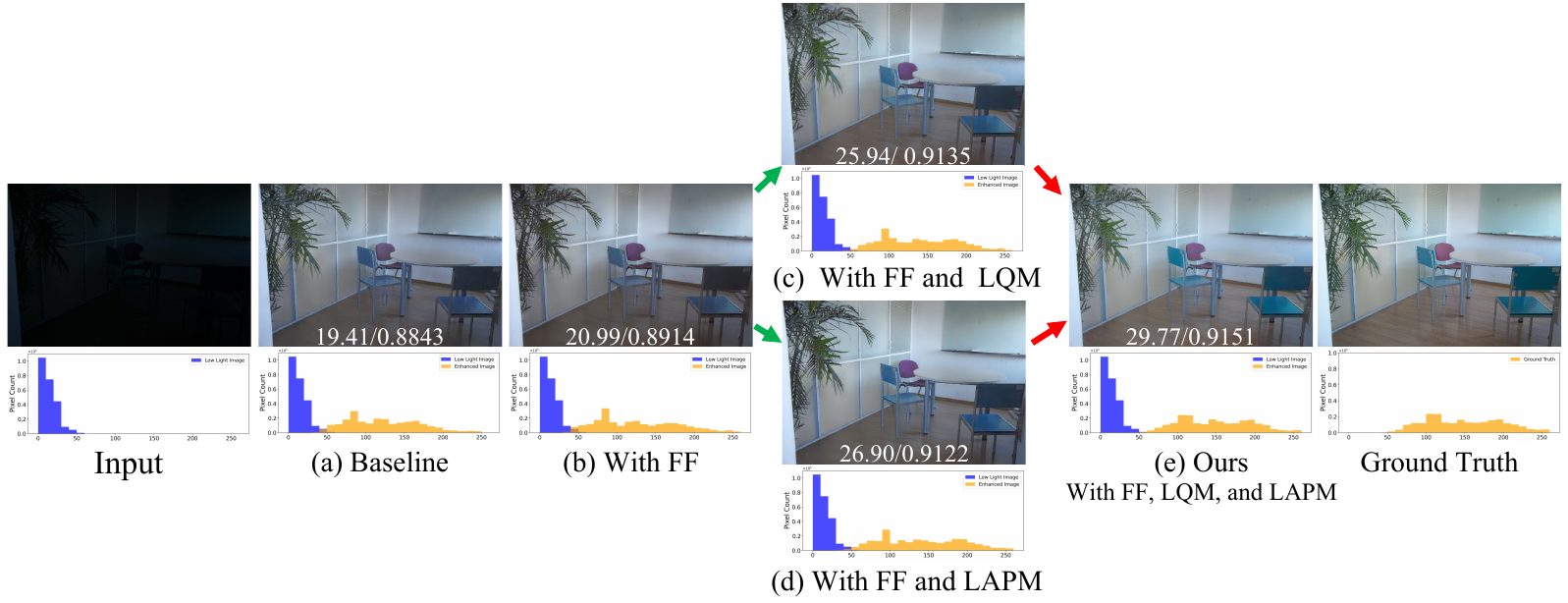}
    \caption{Visual comparison of the ablation studies in Table \ref{table:ablation_structure}.
    Starting from the baseline, the integration of FF, LQM, and LAPM progressively improves both image quality and illumination correction, ultimately achieving the best performance when all components are combined together.}
\label{fig:ablation_structure}
\vspace{-5mm}
\end{figure*}

\subsection{Codes Activation Analysis}
The analysis of code activation frequencies in feature matching provides crucial insights into the efficacy of our image enhancement methods. Fig.~\ref{fig:codebook} (b) shows comparisons of the code activation frequency among baseline, CodeEnhance \cite{xu24}, and the proposed method. From the results, it is observed that directly inputting low-light images into the baseline model ("LL in Baseline") causes highly imbalanced code usage, with certain codes (e.g., index 671) being excessively activated while others are underutilized. In contrast, "LL in CodeEnhance" and "LL in Ours" exhibit more evenly distributed activations, indicating that both methods better mitigate the degradation of code utilization caused by low-light conditions. Notably, the proposed method achieves a closer activation distribution to "GT in Baseline", suggesting that our enhancement strategy more effectively restores diverse and representative feature activations under challenging lighting conditions.

Fig. \ref{fig:Codebook_index_count} illustrates the effectiveness of our method. Panel (a) displays code activation of low-light images, which processed by the proposed method. Panels (b) and (c) compare activation frequencies for our enhanced and ground truth images using a pre-trained VQ-GAN, with panel (d) detailing the top ten code comparisons. 
Panels (a), (c), and (d) collectively demonstrate that our method effectively extracts features from low-light images, achieving activation frequencies that closely match those of the ground truth. This similarity indicates that our method not only improves visibility but also preserves the image's inherent characteristics. Furthermore, comparisons in panels (b), (c), and (d) reveal that the activation patterns of our enhanced images align well with the ground truth within the VQ-GAN, showcasing the high fidelity of our results. 
These results further validate the effectiveness of our method in preserving the integrity and authenticity of image features across diverse lighting conditions.

\begin{table}[!t]
\centering
\caption{Ablation studies of the proposed modules on LOL-v2-real dataset. Baseline is built by VQ-GAN \cite{VQGAN}. FF means the feature fusion $\mathbf{F}_{fuse}$ in skip connection.}
\begin{tabular}{c|cccc|cc}
\toprule
No. & Baseline & FF & LQM & LAPM & PSNR  & SSIM   \\ \midrule
(a)   & \checkmark        &         &      &     & 23.91 & 0.8699 \\
(b)  & \checkmark        & \checkmark       &      &     & 24.82 & 0.8751 \\
(c)   & \checkmark        & \checkmark       & \checkmark    &     & 25.95 & 0.8853 \\
(d)   & \checkmark        & \checkmark       &      & \checkmark   & 27.16 & 0.8911 \\ \midrule
(e)   & \checkmark        & \checkmark       & \checkmark    & \checkmark   & \textbf{28.51} & \textbf{0.8974}\\ \bottomrule
\end{tabular}
\label{table:ablation_structure}
\vspace{-5mm}
\end{table}

\subsection{Ablation Study}
To validate the effectiveness of each proposed component, we conduct ablation studies on the LOL-v2-Real dataset, as summarized in Table~\ref{table:ablation_structure}. The baseline model is constructed based on VQ-GAN \cite{VQGAN}, and we incrementally integrate the feature fusion (FF) in skip connection, Light Quantization Module (LQM), and Light-Aware Prompt Module (LAPM) to systematically assess their contributions.

\textit{1) Study of FF.}
As shown in Table~\ref{table:ablation_structure} and Fig.~\ref{fig:ablation_structure}, integrating the feature fusion module improves the PSNR from 23.91 to 24.82 and the SSIM from 0.8699 to 0.8751.
This performance improvement stems from our learnable linear interpolation mechanism. It uses two dynamically predicted parameters, $\bm{\alpha}$ and $\bm{\beta}$, to modulate decoder features based on encoder information.
Unlike direct feature concatenation or summation, this adaptive interpolation allows the model to enhance fine textures while mitigating noise amplification, thereby achieving more effective reconstruction under low-light conditions.

\begin{figure*}[!t]
    \centering
    \includegraphics[width=0.9 \textwidth]{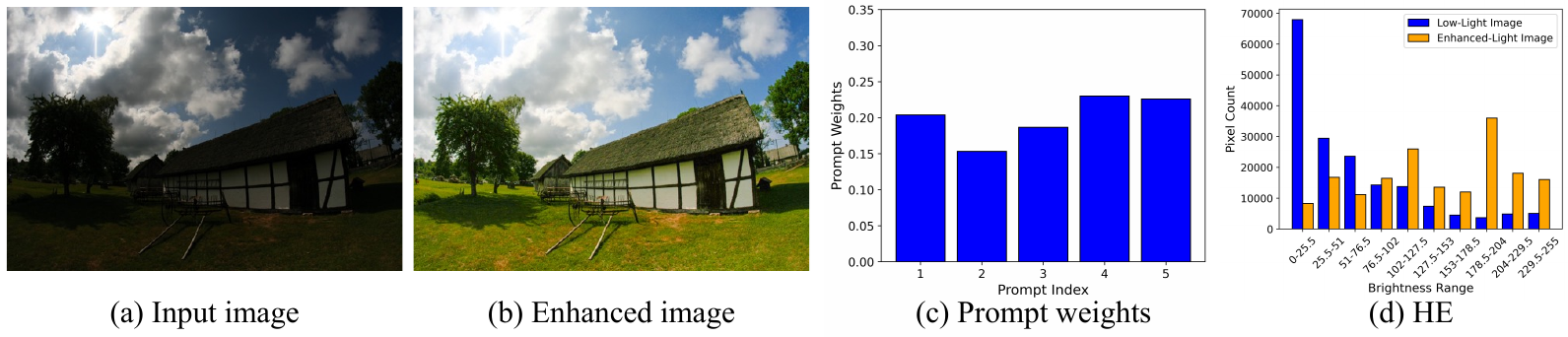}
    \vspace{-4mm}
    \caption{
    Instance analysis of LAPM effects in LLIE.
    (a) and (b) are low-light input image and its enhanced result.
    (c) Learned prompt weights, where prompts 1 and 3 focus on extremely dark regions [0, 25.5], prompt 5 targets transitional brightness [25.5, 51], and prompt 4 responds to mid-to-high brightness levels (above 76.5).
    (d) Brightness distribution histograms before and after enhancement, illustrating luminance correction across different regions.
    }
\label{fig:prompt_MEF}
\vspace{-4mm}
\end{figure*}

\begin{figure}[!t]
    \centering
    \includegraphics[width=0.45 \textwidth]{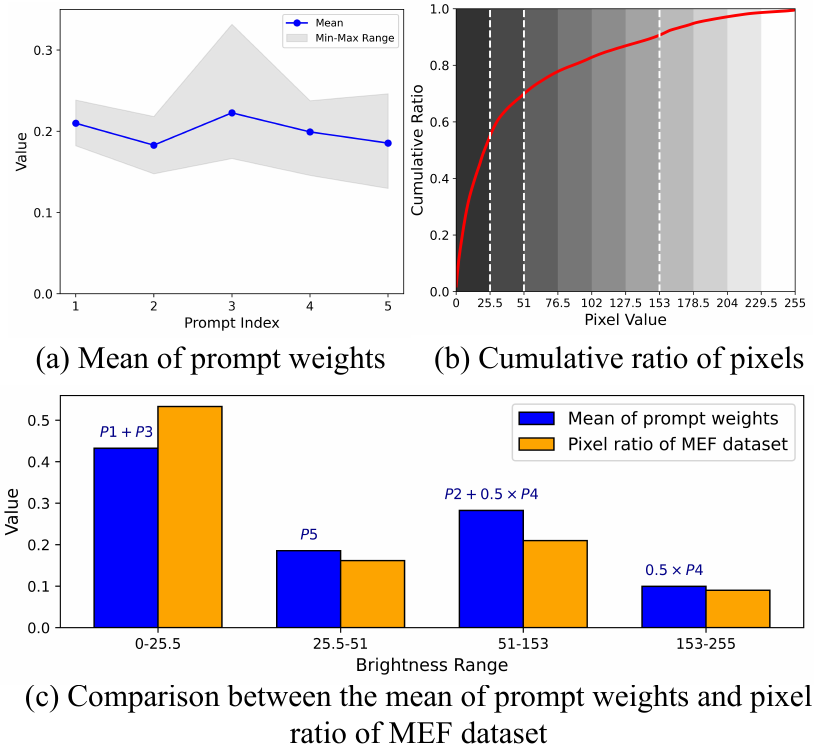}
    \caption{Analysis of learned prompt weights on the MEF dataset \cite{MEF}.
    (a) Mean prompt weights with min–max range for the five learnable prompts.
    (b) Cumulative ratio of pixels as a function of brightness value, with vertical dashed lines marking the boundaries of the five prompt-responsive intervals.
    (c) Comparison of the mean prompt weights against the corresponding pixel ratios within each brightness range, where prompts were grouped according to their correlations in Fig.\ref{fig:prompt_correlation}. [0, 25.5): combined weight of Prompt1$+$Prompt3, [25.5, 51): weight of Prompt5, [51, 153): weight of Prompt2$+0.5\times$Prompt4, [153, 256): weight of $0.5\times$Prompt4. Note that since Prompt4 effectively covers the range [51, 256), we split its weights into two parts, one for [51,153) and one for [153, 256). 
    }
\label{fig:prompt_MEF2}
\vspace{-2mm}
\end{figure}

\textit{2) Study of LQM.}
Upon integrating LQM shown in Table~\ref{table:ablation_structure} and Fig.~\ref{fig:ablation_structure}, the PSNR further improves to 25.95 and the SSIM increases to 0.8853.
This improvement demonstrates the effectiveness of structured illumination modeling through the light-factor space learned by LQM.
By explicitly quantizing illumination-related information and enforcing feature consistency between low-light and normal-light images through the light consistency loss ($\mathcal{L}_{lcl}$) in the learned light-factor space, LQM enables the proposed model to better align feature representations across varying illumination conditions.
This structured alignment process promotes the learning of light-invariant feature representations, significantly enhancing feature robustness and stability, which are crucial for effective low-light image enhancement.

Table~\ref{table:ablation_loss} shows that setting $\lambda$ to 0.5 achieves the best trade-off between luminance consistency and structural detail preservation, yielding the highest image quality and structural fidelity.
A larger $\lambda$ overly constrains illumination consistency, leading to texture loss, while a smaller $\lambda$ weakens illumination-invariance learning.
Thus, $\lambda = 0.5$ allows the model to effectively leverage $\mathcal{L}_{lcl}$, enhancing feature robustness without compromising image quality.

\begin{table}[!t]
\centering
\footnotesize
\caption{Ablation of $\lambda$ for $\mathcal{L}_{lcl}$ on LOL-v2-real dataset.}
\begin{tabular}{c|cccc}
\toprule
 $\lambda$   & 1 & 0.5 & 0.001 \\ \midrule
PSNR &   27.38    & \textbf{28.51}    &   27.43    \\
SSIM &   0.8905  & \textbf{0.8974}    &   0.8928    \\ 
\bottomrule
\end{tabular}
\label{table:ablation_loss}
\vspace{-2mm}
\end{table}

\begin{table}[!t]
\centering
\caption{Ablation of number of prompt vectors on LOL-v2-real dataset.}
\begin{tabular}{c|cc}
\toprule
Number & PSNR    & SSIM   \\ \midrule
3                  & 28.19 & 0.8949 \\
4                  & 28.03 & 0.8911 \\
5                  & \textbf{28.51} & \textbf{0.8974} \\
6                  & 27.47 & 0.8880  \\ \bottomrule
\end{tabular}
\label{table:ablation_prompt_len}
\vspace{-4mm}
\end{table}

\textit{3) Study of LAPM.} 
Based on FF, adding LAPM leads to achieving a PSNR of 27.16 and an SSIM of 0.8911, as shown in Table~\ref{table:ablation_structure}.
LAPM dynamically guides feature learning by injecting illumination-specific prompts, enabling the model to adapt feature representations based on brightness variations.
Compared to the purely static structure offered by LQM, LAPM introduces dynamic flexibility, allowing the model to better respond to complex real-world illumination, thereby yielding significant improvements in image quality shown in Fig.~\ref{fig:ablation_structure}.
Furthermore, Table~\ref{table:ablation_prompt_len} presents the impact of the number of prompt vectors.
We observe that increasing the number of prompts from 3 to 5 leads to continuous improvements in both PSNR and SSIM, reaching the best performance at 5 prompts (28.51 PSNR and 0.8974 SSIM).
However, further increasing the number to 6 results in a noticeable performance drop, due to over-fragmentation of the luminance space, which weakens the effectiveness of each individual prompt.
These results highlight that using 5 prompts achieves the optimal balance between representation capacity and generalization in illumi- nation modeling.

In Fig.~\ref{fig:prompt_MEF}, we first analyze a low-light image whose pixel distribution concentrates within the darkest range ([0, 25.5)). Referring to the correlation analysis from Fig.~\ref{fig:prompt_correlation}, prompts 1 and 3, which are strongly associated with this darkest region, together receive the highest combined weight, effectively enhancing severely underexposed areas. Meanwhile, prompts 4 and 5, assigned relatively lower weights, help refine moderately illuminated regions. 
To further validate this adaptive behavior, we evaluate prompt weights on the entire MEF dataset (see Fig.~\ref{fig:prompt_MEF2}). 
The darkest interval [0, 25.5) contains 53.3\% of pixels and is closely matched by the combined weights of prompts 1 and 3 (0.4324). Transitional low-light pixels ([25.5, 51)) comprise 16.2\%, matched by prompt 5’s weight (0.1854). Mid-range brightness ([51, 153)) accounts for 21.0\%, aligning well with the sum of prompt 2 and half of prompt 4 (0.2825). Finally, the brightest interval ([153, 256)) includes 9.0\% of pixels, corresponding closely with half of prompt 4’s weight (0.0996). Despite slight deviations in exact proportions within these intervals, the learned prompt weights largely reflect the dataset’s overall luminance distribution. 
This analysis confirms that LAPM effectively achieves adaptive illumination-aware feature modulation, allowing the model to selectively emphasize different luminance intervals, thus achieving more natural and visually pleasing results.

\textit{Study of discrete feature number in codebook.}
Table~\ref{table:ablation_feature_num_codebook} presents the impact of the number of discrete features in the codebook on enhancement performance.
We observe that increasing the codebook size from 256 to 1024 progressively improves both PSNR and SSIM, reaching the best performance at 1024 entries.
However, when increased to 2048, it will result in slight degradation, which may be attributed to reduced feature compactness and increase noise sensitivity during reconstruction.
Thus, setting the codebook size to 1024 provides the best trade-off between feature expressiveness and generalization capability.

\begin{table}[!t]
\centering
\caption{Ablation of discrete feature number in codebook on LOL-v2-real dataset.}
\begin{tabular}{c|cc}
\toprule
Number & PSNR    & SSIM   \\ \midrule
256                  & 27.04 & 0.8897 \\
512                  & 27.72 & 0.8946 \\
1024                  & \textbf{28.51} & \textbf{0.8974} \\
2048                  & 27.18 & 0.8923  \\ \bottomrule
\end{tabular}
\label{table:ablation_feature_num_codebook}
\vspace{-3mm}
\end{table}

\begin{figure}[!t]
    \centering
    \includegraphics[width=0.45 \textwidth]{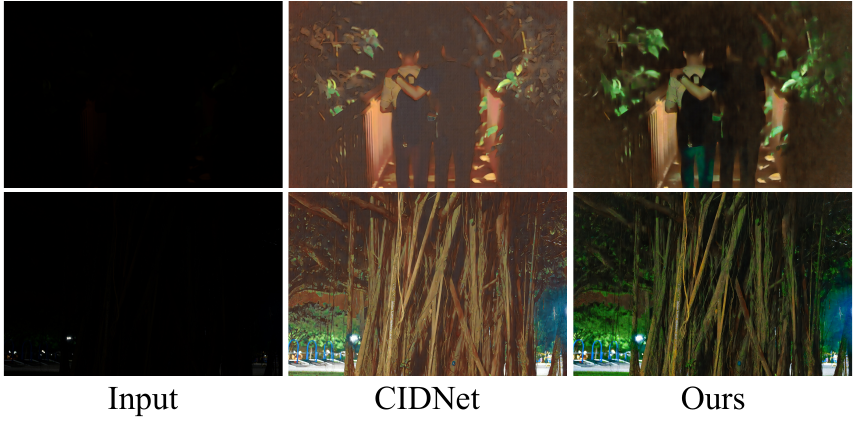}
    \vspace{-4mm}
    \caption{Unsatisfying cases. Images captured using a Sony A7C II camera.
    }
\label{fig:unsatisfying_case}
\vspace{-5mm}
\end{figure}

\subsection{Limitations and future works}
Although LightQANet demonstrates strong performance across diverse lighting conditions, several limitations remain to be addressed.

(1) Extremely dark scenes.
As shown in Fig.~\ref{fig:unsatisfying_case}, LightQANet restores brightness and colors in partially visible regions more effectively than CIDNet, thanks to LQM and LAPM. However, in completely black areas (e.g., the dense canopy), the absence of information causes encoder–codebook misalignment, leading to noise or pseudo-textures. Future work may integrate generative priors to recover plausible structures while suppressing artifacts.
(2) Insufficient high-frequency recovery.
In cross-domain evaluation, fine structures are not always preserved. For example, Fig.~\ref{fig:compare_unpaired} shows blurred license plate characters in the LIME dataset, reflecting limited small-scale detail reconstruction. Future efforts could adopt gradient- or edge-aware losses, leverage high-frequency components via wavelet/Fourier transforms, and utilize classical edge operators (e.g., Sobel, Prewitt, or Canny) to provide explicit priors. In addition, decomposing images into smooth and detail layers with bilateral filtering, selectively reinforcing the detail layer, and applying lightweight sharpening as post-processing may further refine structural fidelity, collectively mitigating edge blurring and detail loss.

\section{Conclusion} \label{conclusion}
In this study, we propose a novel LightQANet framework for LLIE, which emphasizes light-invariant feature learning through both structured quantization and dynamic adaptation.
Specifically, we design an LQM to extract and quantize light-relevant information within feature representations, thereby effectively bridging the gap between low-light and normal-light conditions, so as to promote the learning of light-invariant features.
In addition, we introduce an LAPM that dynamically encodes illumination priors to adaptively guide feature learning across varying brightness levels.
Extensive experiments across multiple datasets, including both same-source and cross-source scenarios, demonstrate that LightQANet consistently outperforms the existing state-of-the-art LLIE methods, validating the effectiveness of our proposed approach in achieving robust and adaptive illumination enhancement.

\section{Acknowledgment}
This work was supported in part by National Natural Science Foundation of China (No. 62476172, 62476175, 62272319, 62206180, 82261138629), and Guangdong Basic and Applied Basic Research Foundation (No. 2023A1515010677, 2023B1212060076, 2024A1515011637, 2025A1515011511), and Science and Technology Planning Project of Shenzhen Municipality (No. JCYJ20220818095803007, JCYJ20240813142206009), and Guangdong Provincial Key Laboratory (No. 2023B1212060076), and XJTLU Research Development Funds (No. RDF-23-01-053).

\ifCLASSOPTIONcaptionsoff
  \newpage
\fi

\bibliographystyle{./bibliography/IEEEtran}
\bibliography{reference.bib}

\end{document}